\documentclass[10pt]{article}

\newif\ifdraft
\draftfalse
\ifdraft
\usepackage{lineno}
\linenumbers
\fi

\usepackage[utf8]{inputenc} % allow utf-8 input
\usepackage[T1]{fontenc}    % use 8-bit T1 fonts
\usepackage{lmodern}
\usepackage{amsthm}
\usepackage{amsmath}
\usepackage{amssymb}
\usepackage{hyperref}       % hyperlinks
\usepackage{url}            % simple URL typesetting
\usepackage{booktabs}       % professional-quality tables
\usepackage{amsfonts}       % blackboard math symbols
\usepackage{mathtools}      % useful tools for mathematical typesetting
\usepackage{nicefrac}       % compact symbols for 1/2, etc.
\usepackage{microtype}      % microtypography
\usepackage{subfiles}       % loading tex files
\usepackage{bbold}          % for \mathbb{1} to work
\usepackage{xcolor}         % enable coloured text
\usepackage{bm}             % for bold math symbols
\usepackage[font=footnotesize]{subcaption}     % for side by side figures
\usepackage[font=footnotesize]{caption}
\usepackage{enumitem}
\usepackage[numbers, compress]{natbib}
\usepackage[verbose=true,letterpaper]{geometry}
\usepackage[parfill]{parskip}

\mathtoolsset{showonlyrefs}

\AtBeginDocument{
  \newgeometry{
    textheight=9in,
    textwidth=5.5in,
    top=1in,
    headheight=12pt,
    headsep=25pt,
    footskip=30pt
  }
}

\begingroup
\makeatletter
\@for\theoremstyle:=definition,remark,plain\do{%
  \expandafter\g@addto@macro\csname
  th@\theoremstyle\endcsname{%
    \addtolength\thm@preskip\parskip
  }%
}
\endgroup

\definecolor{myred}{HTML}{880000}
\definecolor{mygreen}{HTML}{008800}
\definecolor{myblue}{HTML}{000088}
\definecolor{linkblue}{HTML}{0000BB}

\hypersetup{
  colorlinks,
  linkcolor={myred},
  citecolor={myblue},
  urlcolor={linkblue}
}

\newtheorem{lemma}{Lemma}

\newtheorem{theorem}{Theorem}

\newtheorem{definition}{Definition}

\renewcommand{\th}{\ensuremath{^\mathit{th}}}
\usepackage{mymacros}

\title{The Statistical Complexity of Early-Stopped Mirror Descent
\ifdraft
\footnote{A version of this work is currenly under review for the NeurIPS 2020
conference.}
\fi
}

\author{
  Tomas Va\v{s}kevi\v{c}ius\textsuperscript{1},
  Varun Kanade\textsuperscript{2},
  Patrick Rebeschini\textsuperscript{1} \\
  \textsuperscript{1} Department of Statistics,
  \textsuperscript{2} Department of Computer Science \\
  University of Oxford \\
  \texttt{\{tomas.vaskevicius,
  patrick.rebeschini\}@stats.ox.ac.uk} \\
  \texttt{varunk@cs.ox.ac.uk}
}

\begin{document}

\maketitle

\begin{abstract}%

Recently there has been a surge of interest in understanding implicit
regularization properties of iterative gradient-based optimization algorithms.
In this paper, we study the statistical guarantees on the excess risk achieved
by early-stopped unconstrained mirror descent algorithms applied to the
unregularized empirical risk with the squared loss for linear models and kernel
methods. By completing an inequality that characterizes convexity for the
squared loss, we identify an intrinsic link between offset Rademacher
complexities and potential-based convergence analysis of mirror descent
methods. Our observation immediately yields excess risk guarantees for the path
traced by the iterates of mirror descent in terms of offset complexities of
certain function classes depending only on the choice of the mirror map,
initialization point, step-size, and the number of iterations.  We apply our
theory to recover, in a clean and elegant manner via rather short proofs, some
of the recent results in the implicit regularization literature, while also
showing how to improve upon them in some settings.

\end{abstract}

%--------------------------------- Sub-files ----------------------------------
%------------------------------------------------------------------------------

\section{Introduction}
\label{section:introduction}

In a typical statistical learning setup, we observe a dataset $D_{n}$ of $n$
input-output pairs $(x_{i}, y_{i}) \in \R^{d} \times \R$ sampled i.i.d.\ from
some unknown distribution $P$.
When learning with respect to the quadratic loss, the goal is to output a function
$\widehat{g} = \widehat{g}(D_{n}) : \R^{d} \rightarrow \R$ which minimizes the
\emph{risk} $R({\widehat{g}})$ defined as follows, for any square-integrable
function $g$:
\begin{equation*}
	R(g) = \exd{(X,Y)\sim P}{\left(g(X) - Y\right)^{2}}.
\end{equation*}
Among the most studied statistical estimators is the \emph{empirical risk
  minimization} (ERM) algorithm, which given a function class $\mathcal{G}$
outputs a function $\widehat{g}_{\mathcal{G}} = \widehat{g}_{\mathcal{G}}(D_{n})$
defined as
\begin{equation}
  \label{eq:erm-definition}
  \widehat{g}_{\mathcal{G}} \in
  \underset{{g \in \mathcal{G}}}{\operatorname{arg\, min}}
  \,R_{n}(g),
  \enskip\text{ where }\enskip
  R_{n}(g) := \frac{1}{n}\sum_{i=1}^{n}(g(x_{i}) - y_{i})^{2},
  %\enskip\text{ for all } g,
\end{equation}
in some cases with a regularization penalty term added to the optimization
objective $R_{n}(g)$, such as $\ell_{p}$ norm of the model parameters.
We consider the \emph{non-realizable} or
\emph{agnostic} setting, i.e., the case in which there is no assumption that $\exd{}{Y | X}$ is
determined by a well-specified model from a reference class of functions.
In the agnostic case, a key performance
measure of an estimator $\widehat{g}$ is its $\emph{excess risk}$ with respect
to some reference class of functions $\mathcal{F}$:
\begin{equation*}
  \mathcal{E}(\widehat{g}, \mathcal{F})
  = R(\widehat{g}) - \inf_{f \in \mathcal{F}} R(f).
\end{equation*}

\begin{figure}[ht]
  \centering
  \begin{subfigure}{.45\textwidth}
    \centering
    \includegraphics[width=\linewidth]{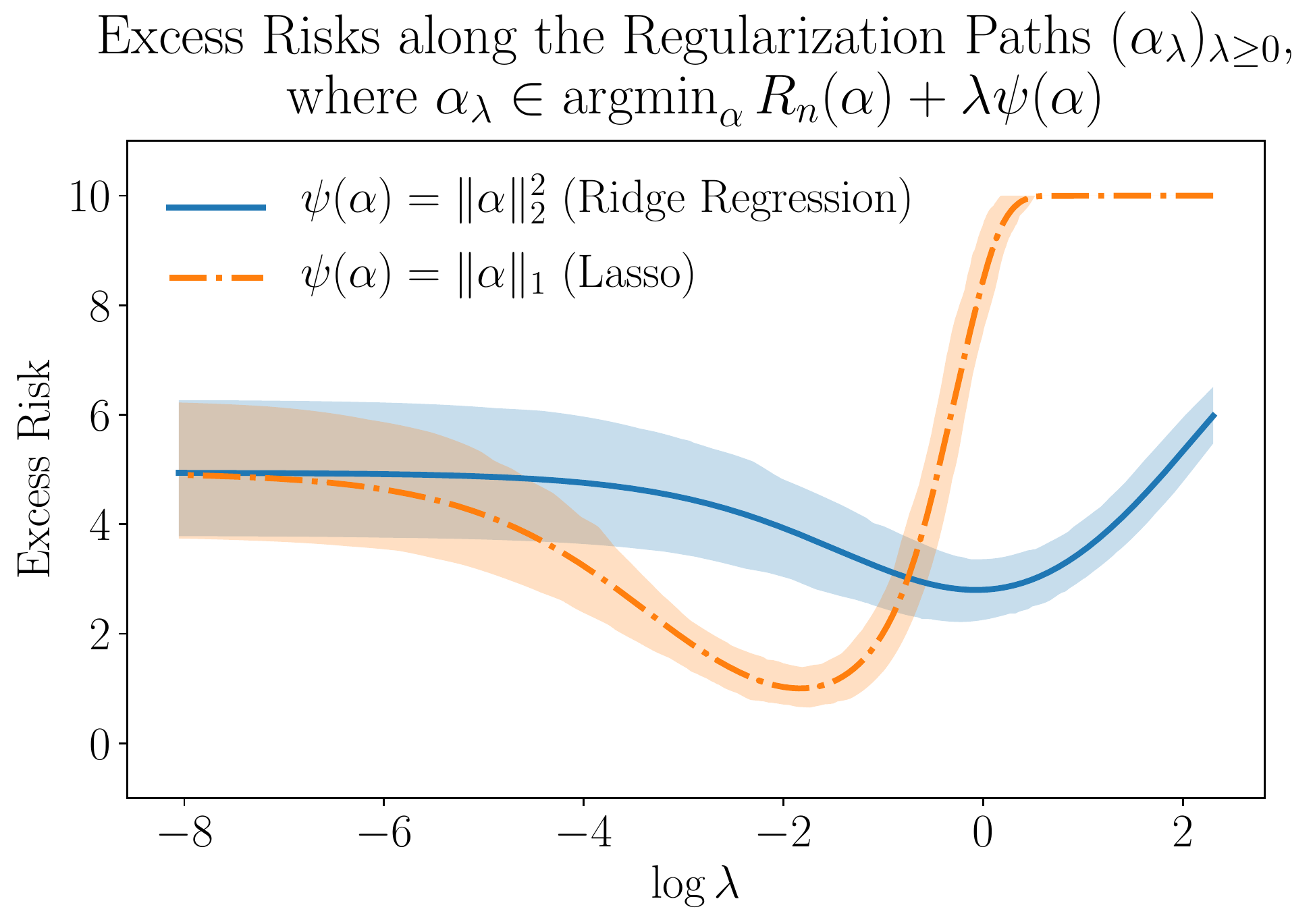}
    \caption{Explicit regularization.}
    \label{fig:explicit-regularization}
  \end{subfigure}%
  \begin{subfigure}{.45\textwidth}
    \centering
    \includegraphics[width=\linewidth]{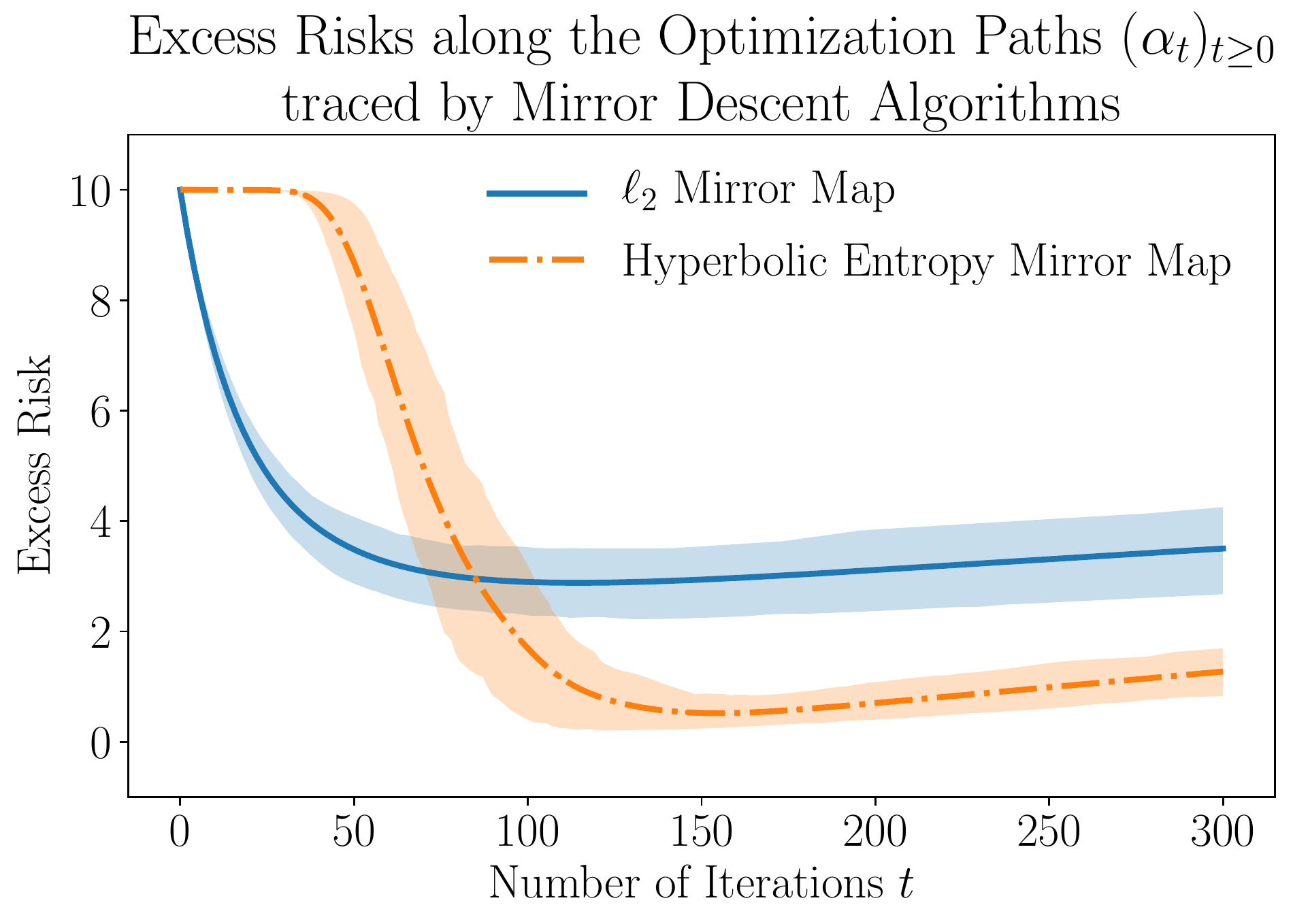}
    \caption{Implicit regularization.}
    \label{fig:implicit-regularization}
  \end{subfigure}%
  \caption{Consider a distribution $P$ such that
  $X \sim N(0, I_{d})$ and $Y \vert X = x \sim \ip{\alpha'}{x} + N(0, 5^{2})$
  for some parameter $\alpha' \in \R^{d}$.
  Fix $n = 200, d=100$ and let $\alpha'$ be a $10$-sparse vector with non-zero
  entries equal to $\pm 1$.
  Due to the sparsity of $\alpha'$, explicit regularization via $\ell_{1}$
  penalization results in a class of models $(\alpha_{\lambda})_{\lambda \geq 0}$
  that at its minimum achieves significantly lower risk than the
  class of models generated via $\ell_{2}$ penalization
  (cf.\ Figure~\ref{fig:explicit-regularization}).
  Figure~\ref{fig:implicit-regularization} demonstrates a similar phenomenon
  from an implicit regularization point of view. Due to the
  sparsity of $\alpha'$, the choice of a hyperbolic entropy mirror map
  (cf.\ Section~\ref{section:application-l1}) yields an optimization path
  that at its minimum achieves excess risk nearly an order of magnitude lower than
  the path generated by the vanilla gradient descent updates.
  In the plot above, the solid lines denote means over $100$ runs whereas the shaded regions
  correspond to the $10\th$ and the $90\th$ percentiles.}
  \label{fig:explicit-vs-implicit-regularization}
\end{figure}

Traditionally, in learning theory, statistical and computational
properties of ERM estimators have been considered separately.
From a statistical point of view, localized complexity measures
have become a default tool in statistical learning theory and
empirical processes theory for controlling the excess
risk of ERM algorithms $\widehat{g}_{\mathcal{G}}$
with respect to the function class $\mathcal{G}$ itself, i.e., for controlling
$\mathcal{E}(\widehat{g}_{\mathcal{G}}, \mathcal{G})$~\citep{bartlett2005local, koltchinskii2006local}.
A rich and general theory regarding these complexity measures has been
developed and used to provide excess risk bounds in both classification and
regression settings, yielding minimax-optimal results in several cases.  Such
complexity measures depend on combinatorial or geometric parameters of
interest, such as the VC-dimension or eigenvalue decay of the kernel matrix
and, in particular, they serve as a guiding principle to choose a suitable
\emph{explicit regularizer} for a set of candidate models
$(\widehat{g}_{\mathcal{G}_{\lambda}})_{\lambda \in
\Lambda}$, where $\lambda \in \Lambda$ is a hyper-parameter that controls
the amount of regularization.
In practice, some $\lambda^{\star} \in \Lambda$ is then chosen via
some model selection procedure such as cross-validation, aiming
to select a model with the smallest risk.
From a computational point of view, computing the estimators
$(\widehat{g}_{\mathcal{G}_{\lambda}})_{\lambda \in \Lambda}$ can be
done by solving the corresponding optimization problems defined in
Equation~\eqref{eq:erm-definition}, one for each $\lambda \in \Lambda$.
An appealing aspect of this approach is
that the design and analysis of efficient optimization algorithms, exploiting
the geometry of $G_\lambda$ that arises from the the structure of the model as
well as the distribution $P$, can be done independently of the statistical
analysis of its performance.

Recent years have also witnessed an increased interest in directly studying the
statistical properties of models trained by gradient-based methods,
particularly in relation to the notions of \emph{implicit regularization} and
\emph{early stopping}. For a family of functions $\mathcal{G} = \{g_{\alpha} :
\alpha \in \R^{m}\}$ parametrized by a vector $\alpha$, such methods are fully
characterized by the initialization point $\alpha_{0}$ and an update rule,
which given $\alpha_{t}$ and the gradient of the empirical risk at
$\alpha_{t}$, generates the next iterate $\alpha_{t+1}$, yielding a set of
candidate estimators $(\widehat{g}_{\alpha_{t}})_{t\geq0}$. Early stopping has
an effect
akin to \emph{explicit} regularization discussed above, and the \emph{stopping
time} $t^\star$ can be chosen in practice via cross-validation, just as in
the case of choosing the explicit regularization parameter $\lambda^{\star}$
corresponding to the best model among
$(\widehat{g}_{\mathcal{G}_{\lambda}})_{\lambda \in \Lambda}$.  In modern
large-scale machine learning applications, early stopping is often the preferred way to
perform model selection, since obtaining a new model is as cheap as
performing a step of
gradient descent, as opposed to solving a new optimization
problem with a different regularization parameter.
In Figure~\ref{fig:explicit-vs-implicit-regularization}, we demonstrate that
different choices of optimization algorithms applied to the unregularized
empirical risk $R_{n}$ yield different statistical performance along
the optimization path $(\widehat{g}_{\alpha_{t}})_{t \geq 0}$,
in a similar way that a choice of an explicit regularizer
affect the statistical performance along the corresponding regularization path.

It is by now well understood that changing the update rule that
generates the sequence $(\widehat{g}_{\alpha_{t}})_{t\geq0}$, e.g., by changing
the optimization algorithm or parametrization of the model class, can directly
affect both the statistical properties of the iterates $\widehat{g}_{\alpha_{t}}$,
as well as computational properties, such as an upper-bound on the optimal
stopping time $t^{\star}$. However,
most of the literature has focused on the investigation of
vanilla gradient descent updates:
$\alpha_{t+1} = \alpha_{t} - \eta \nabla_{\alpha_{t}}R_{n}(\widehat{g}_{\alpha_{t}})$
(cf.\ Section~\ref{section:related-work}).
The existing theory does not easily generalize to other update rules
corresponding to different problem geometries.
A general theory that connects the notion of early stopping for a more general
class of update rules with the well-established theory of
localized complexities is still missing.
More broadly, a general ``language'' to reason
about the statistical properties of trajectories traced by optimization algorithms
applied to the unregularized empirical risk is still lacking.

In this paper, we study a \emph{family} of update rules given by the mirror descent algorithm
\citep{nemirovsky1983problem,beck2003mirror}.  Mirror descent, which includes vanilla gradient descent as a special case, is increasingly becoming the tool of choice in optimization
and machine learning, applied well beyond the traditional settings of
convex optimization and
online learning. Among the properties that make mirror descent appealing are
its ability to exploit non-Euclidean geometries via properly designed mirror
maps, the fact that the algorithm admits a general potential-based convergence
analysis in terms of Bregman divergences, and its ability to represent
a large class of algorithms in a unified and well-developed framework.

We consider a setting where conditionally on the observed data $D_{n}$
there exists a matrix $Z \in \R^{n \times m}$ such that
the parametric family of functions $\{g_{\alpha} : \alpha \in \R^{m}\}$
satisfies $g_{\alpha}(x_{i}) = (Z\alpha)_{i}$ for all $i = 1, \dots, n$.
As special cases, our setup admits linear regression and kernel methods
(cf.\ Section~\ref{section:applications}), cornerstones of modern statistics and
machine learning.
Our work reveals an inherent connection between the statistical properties
of the mirror descent iterates $(\widehat{g}_{\alpha_{t}})_{t\geq0}$ and
the notion of offset Rademacher complexity \citep{liang2015learning}.
Consequently, our work unearths a simple and elegant way to simultaneously
analyze upper-bounds on the stopping time $t^{\star}$, as well as the excess risk
$\mathcal{E}(\widehat{g}_{\alpha_{t}}, \mathcal{F})$ for all $t \leq t^{\star}$
in terms of the mirror map, the initialization point $\alpha_{0}$, the step-size,
and the function class $\mathcal{F}$. Through a simple one page analysis, we
are able to rederive (nearly identical) results from prior work connecting
early stopping and (optimal) statistical performance that previously
involved several pages of low-level arguments.%
\footnote{
  In some cases, the results we obtain are not \emph{exactly}
  comparable to the ones obtained in the related work, as some
of our assumptions are considerably weaker, e.g.\ \emph{non-realizable} setting,
and our guarantees stronger, excess risk vs bounds in
$\inlinenorm{\cdot}_{n}^{2}$ or $\inlinenorm{\cdot}_{P}^{2}$ norms defined
in Section~\ref{section:background}.
However, in some
applications we require boundedness which is not required in some of the prior work,
and some of our results are stated only in expectation, rather than high
probability. We note that we can also easily obtain high-probability results in some
settings (e.g.\ heavy-tailed classes under the lower-isometry assumption)
that are outside the scope of the related work in the early stopping
literature. See Section~\ref{section:application-nonparametric-regression}
for an extended discussion.}
Additionally, in the well-studied case of Euclidean gradient descent, our work
improves upon the prior results connecting early stopping to localized
complexity measures \citep{raskutti2014early, wei2019early} by providing
upper-bounds on the expected excess risk without any distributional assumptions
on $P$ other than boundedness (cf.\ Section~\ref{section:related-work}).

\subsection{Background}
\label{section:background}

We begin this section by explaining the difficulties involved in analyzing
early-stopped iterative algorithms via the classical notion of localized
complexities.  We then describe the offset Rademacher complexities, which is
a form of localization based on a different mathematical machinery that is more
suitable for our setting. Finally, we define the mirror descent updates and
outline a short well-known potential-based proof of its convergence.

In what follows, we let $\Vert g - f\Vert_n^{2} =
\frac{1}{n}\sum_{i=1}^{n}(g(x_{i}) - f(x_{i}))^{2}$ and $\Vert g - f
\Vert_P^{2} = \ex{(g(X) - f(X))^{2}}$ denote the empirical and population
$\ell_{2}$ distances between functions $g$ and $f$, respectively.
Further, given a function class $\mathcal{F}$, we denote by
$g_{\mathcal{F}} \in \mathcal{F}$ a function
that attains risk equal to $\inf_{g \in \mathcal{F}} R(g)$.\footnote{
  If such a function $g_{\mathcal{F}}$ does not exist, we can redefine $g_{\mathcal{F}}$ to be any
function in $\mathcal{F}$ such that $R(g_{\mathcal{F}}) \leq \inf_{g \in
\mathcal{F}} R(g) + \delta$ for any arbitrarily small $\delta > 0$.
}
A table of notation is provided in Appendix~\ref{appendix:table-of-notation}.
%------------------ Violation of Bernstein Condition Figure -------------------
%------------------------------------------------------------------------------
\begin{figure}[ht]
  \centering
  \includegraphics[width=0.9\linewidth]{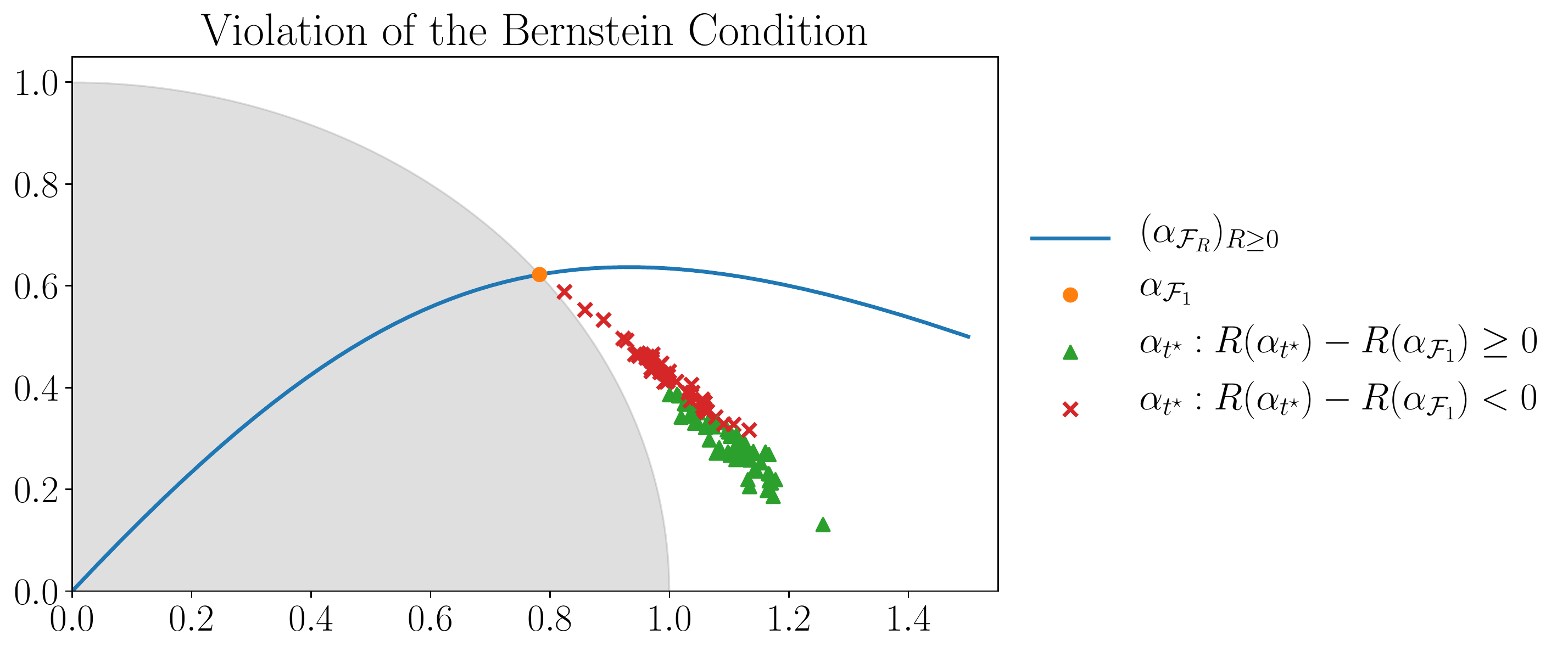}
  \captionsetup{singlelinecheck=false} % Allows display math in caption.
  \caption[singlelinecheck=false]{
    Fix $\alpha' = (1.5, 0.5)^{\mathsf{T}}$, $n=100$
    and consider a distribution $P$ defined as
    \begin{equation*}
      X \sim N(0, \Sigma)\text{ and }
      Y \vert X = x \sim \langle \alpha', x \rangle + N(0, 0.5^{2}),
      \text{ where }
      \Sigma = \begin{bmatrix}1 & 1 \\ 1 & 2\end{bmatrix}.
    \end{equation*}
    For simplicity, we denote linear functions $\ip{\alpha}{\cdot}$
    by the parameter $\alpha$. For any $R \geq 0$ let
    $\mathcal{F}_{R} = \{\alpha : \norm{\alpha}_{2} \leq R\}$ denote an
    $\ell_{2}$ ball of radius $R$ and let
    $\alpha_{\mathcal{F}_{R}} = \operatorname{argmin}_{\alpha \in \mathcal{F}_{R}} R(\alpha)$
    denote the population risk minimizer in $\mathcal{F}_{R}$.
    Let $\widehat{g}$ be some estimator and suppose that we want to upper-bound
    $\mathcal{E}(\widehat{g}, \mathcal{F}_{1})$, where the function class
    $\mathcal{F}_{1}$ is denoted by the shaded region in the plot above.
    The theory of localized Rademacher complexities can be readily used to
    upper-bound $\mathcal{E}(\widehat{g}, \mathcal{F}_{1})$ for
    proper algorithms (i.e., estimators $\widehat{g} \in \mathcal{F}_{1}$)
    that with probability $1$ satisfy
    $R_{n}(\widehat{g}) \leq R_{n}(\alpha_{\mathcal{F}_{1}})$, for instance,
    if $\widehat{g}$ is an ERM algorithm over the class $\mathcal{F}_{1}$.
    The classical theory of localization is, on the other hand, less suitable
    for analysis of unconstrained iterative algorithms.
    To see why, consider running mirror descent with the mirror map
    $\psi(\alpha) = \alpha^{\mathsf{T}}\Sigma\alpha/2$, the initialization $\alpha_{0} = 0$
    and the step-size $\eta = 10^{-3}$.
    Define the stopping time $t^{\star} = \min
    \{\,t \geq 0 : R_{n}(\alpha_{t}) \leq R_{n}(\alpha_{\mathcal{F}_{1}})\}$
    so that our early-stopped estimator is identified with the parameter $\alpha_{t^{\star}}$.
    We plot the values of $\alpha_{t^{\star}}$ (denoted by crosses and triangles)
    over $100$ runs, where the crosses denote instances of $\alpha_{t^{\star}}$ such
    that $R(\alpha_{t^{\star}}) < R(\alpha_{\mathcal{F}_{1}})$.
    Such points, in particular, violate the Bernstein condition for all $C > 0$
    (i.e., it does not hold that $R(\alpha_{t^{\star}}) -
      R(\alpha_{\mathcal{F}_{1}}) \geq
      C\inlinenorm{\alpha_{t^{\star}} - \alpha_{\mathcal{F}_{1}}}_{P}^{2}$)
    and hence demonstrate that statistical analysis of early-stopped mirror
    descent estimators does not easily fit within the classical framework of
    localized complexity measures.
    In contrast, all points in $\mathcal{F}_{1}$ (denoted by the shaded ball)
    satisfy the Bernstein condition with parameter $C=1$; thence, bounds
    on $\mathcal{E}(\widehat{g}, \mathcal{F}_{1})$ can be easily obtained
    via the classical notion of localization whenever $\widehat{g}$
    is a proper estimator (i.e., $\widehat{g} \in \mathcal{F}_{1}$) such that
    $R_{n}(\widehat{g}) \leq R_{n}(\alpha_{\mathcal{F}_{1}})$ almost surely.
  }
  \label{fig:violation-of-bernstein-condition}
\end{figure}
\vspace{-1.5em}
\paragraph{Localized Rademacher complexities.}
The classical notion of global Rademacher complexities
\cite{bartlett2002rademacher} can only establish
the slow rates of order $n^{-1/2}$ on the excess risk (cf.\
\cite[Theorem 2.3]{bartlett2006empirical}). This observation was one of the primary
motivating factors in the development of localized Rademacher complexities
\cite{bartlett2005local, koltchinskii2006local}.
Let $\mathcal{G}$ be the range of an estimator $\widehat{g}$.
Rather than considering the Rademacher complexity of the whole function
class $\mathcal{G}$, localization builds on the idea of computing the
Rademacher complexity of the smaller class
$\{g \in \mathcal{G} : \Vert g - g_{\mathcal{G}}\Vert_n^{2} \leq r\}$ for some
suitably defined radius $r$ that can be obtained by solving a certain
fixed-point equation. More recent work focuses on unbounded and, in particular,
heavy-tailed settings \cite{mendelson2014learning} as well as extending the
scope of localization to study estimators other than ERM, e.g.,
to study the statistical performance of tournament
procedures \cite{lugosi2016risk, mendelson2017extending}.
Crucially, this line of research is rooted in the following two assumptions.
First, $(P, \mathcal{G})$ is assumed to satisfy a convexity type assumption
known in the literature as the
\emph{Bernstein condition} (cf.\ \cite{bartlett2006empirical}), which states that for some
constant $C > 0$, $R(g) - R(g_{\mathcal{G}}) \geq C \Vert g - g_{\mathcal{G}}\Vert_P^{2}$
for any $g \in \mathcal{G}$.
If the class
$\mathcal{G}$ is convex and the loss function is quadratic then this condition
follows immediately by convexity with $C=1$ (see \cite[Definition 5.2]{mendelson2017extending}
for more details).
The second condition is imposed on the estimator $\widehat{g}$ itself (rather
than its range $\mathcal{G}$), which
requires that the inequality
$R_{n}(\widehat{g}) \leq R_{n}(g_{\mathcal{G}})$ holds for all
realizations of $D_{n}$,
a property naturally satisfied by the ERM algorithm over the class
$\mathcal{G}$. Our setting, however, does not easily fit into the above
assumptions. To see why, note that the sequence
$(\widehat{g}_{\alpha_{t}})_{t\geq0}$ obtained by some iterative
algorithm aimed at minimizing the unconstrained empirical risk
is not necessarily explicitly constrained to lie in the class
$\mathcal{G}$. Thus by the time the inequality
$R_{n}(\widehat{g}_{\alpha_{t}}) \leq
R_{n}(g_{\mathcal{G}})$ is satisfied, the iterate
$\widehat{g}_{\alpha_{t}}$ can already be outside the class $\mathcal{G}$,
potentially violating the Bernstein condition (cf.\
Figure~\ref{fig:violation-of-bernstein-condition}) in all cases except when
$\mathcal{G}$ is taken to be the union of ranges of $\widehat{g}_{\alpha_{t}}$
over all $t \geq 0$.

%---------------------------- Offset Complexities -----------------------------
%------------------------------------------------------------------------------
\paragraph{Offset Rademacher complexities.}
When learning with the quadratic loss, a theory of localization
based on shifted Rademacher processes was developed by
\citet{liang2015learning} inspired by prior work in online
learning \cite{rakhlin2014online}.
The use of shifted empirical processes in order to bypass technicalities
present in the classical localization arguments date back at least to
\citep{wegkamp2003model} and have recently found applications in
cross-validation \citep{lecue2012oracle}, classification \citep{zhivotovskiy2018localization}
and PAC-Bayes bounds \citep{yang2019fast}.
For a function class $\mathcal{G}$, a dataset $D_{n}$, an independent sequence
of Rademacher random variables $\sigma_{1}, \dots, \sigma_{n}$, and any
$c \geq 0$, the \emph{empirical offset Rademacher complexity} is defined as,
conditionally on the observed data $D_{n}$:
\begin{equation}
    \label{eq:offset-complexity-definition}
    \mathfrak{R}_{n}(\mathcal{G}, c) =
    \exd{\sigma_{1}, \dots, \sigma_{n}}{
    \sup_{g \in \mathcal{G}}
    \left\{
		 \frac{1}{n}\sum_{i=1}^{n}(2\sigma_{i}g(x_{i})
		 - cg(x_i)^2)
    \right\}
      }.
\end{equation}
Note that since the terms $-cg(x_i)^2$ are always non-positive, the above notion
of complexity is never larger than global Rademacher
complexity of the class $\mathcal{G}$, which is recovered with $c=0$.
On the other hand, for any $c>0$, the quadratic term in the above definition
has a localization effect by compensating for the fluctuations in the term
involving Rademacher variables (see Section 5.2 and the discussion following
Theorem 3 in \citep{liang2015learning}).
Importantly, the theory of localization via offset complexities replaces
the Bernstein condition
used in the classical theory of localization by
the \emph{offset condition} defined below.
\begin{definition}[Offset condition]
  \label{definition:offset-condition}
  A triple $(P, \mathcal{F}, \widehat{g})$ satisfies the offset condition with
  parameters $\varepsilon \geq 0, c > 0$, if for $D_n \sim P^n$, with probabilty $1$, we have
  $ R_{n}(\widehat{g}) - R_{n}(g_{\mathcal{F}})
    + c\Vert \widehat{g} - g_{\mathcal{F}}\Vert_n^{2} \leq \varepsilon.$
\end{definition}
The above condition with $\varepsilon = 0$ was introduced in
\citep{liang2015learning} where it was called the \emph{geometric inequality} and
shown to hold for ERM estimators over convex classes $\mathcal{F}$ as well as
the two-step star estimator \citep{audibert2008progressive} over general
classes for finite aggregation.\footnote{
  We show in Appendix~\ref{appendix:on-epsilon-term} that
  the $\varepsilon$ term affects the resulting excess risk bounds only by an
  additive term equal to $\varepsilon$, which can be chosen to be arbitrarily
  small in our main results.
} A key advantage offered by the theory of offset
complexities is that the range of $\widehat{g}$ need not be a subset of
$\mathcal{F}$, as long as the offset condition is satisfied. This allows us to consider very general estimators $\widehat{g}$, possibly with non-convex
ranges $\mathcal{G}$. In this respect, our work
can be seen as showing that early-stopped mirror descent satisfies the offset
condition defined above. Once an estimator is shown to satisfy the offset
condition, its excess risk $\mathcal{E}(\widehat{g}, \mathcal{F})$ can be
controlled in terms of the offset complexity
$\mathfrak{R}_{n}(\mathcal{G} - g_{\mathcal{F}}, c)$.
The theory developed in \cite{liang2015learning} establishes high-probability
bounds under the lower-isometry assumption, which can hold even for possibly
heavy-tailed classes (\citep[Theorem 4]{liang2015learning}), as well as bounds
in expectation under no assumptions other than boundedness (\citep[Theorem
3]{liang2015learning}). The result in expectation states that given
$\sup_{g \in \mathcal{G} \cup g_{\mathcal{F}}} \inlineabs{g}_{\infty} \leq B$
and $\inlinenorm{Y}_{L_{\infty}(P)} \leq M$ for some $B, M > 0$,
we have
\begin{equation}
  \label{eq:bound-in-expectation-no-assumptions}
  \ex{\mathcal{E}(\widehat{g}, \mathcal{F})} \leq c_{1}\ex{
    \mathfrak{R}_{n}(\mathcal{G} - g_{\mathcal{F}}, c_{2})
  } + \varepsilon,
\end{equation}
where $c_{1} = (4 + c/2)B + 2M$ and $c_{2} = c/(4(B+M)(2 + c))$ and the
expectation is taken over datasets $D_{n}$; here $c$ and $\varepsilon$ are those that appear in the offset condition.
The generality of the above result allows us to improve upon the existing
bounds in the early stopping literature even for gradient descent updates
(cf.\ Section~\ref{section:related-work}).
%------------------------------- Mirror Descent -------------------------------
%------------------------------------------------------------------------------
\paragraph{Mirror descent.}
The key object characterizing the geometry of the mirror descent algorithm is
the \emph{mirror map} $\psi$, a strictly convex and differentiable
function mapping some open set $\mathcal{D} \subseteq \R^{m}$ to $\R$
whose gradient is surjective, i.e.\
$\{\nabla \psi(\alpha) \mid \alpha \in \mathcal{D} \} = \R^{m}$.
By slightly abusing notation, we use $R_n(\alpha) := R_n(g_\alpha)$ to
denote the empirical risk of $g_\alpha$. When optimizing the empirical risk
$R_n(\alpha)$, the mirror descent updates in continuous and discrete time
are given respectively by
\begin{equation}
  \label{eq:mirror descent-updates}
  \frac{d}{dt} \alpha_{t} = -\left(\nabla^{2}
  \psi(\alpha_{t})\right)^{-1}\nabla R_n(\alpha_{t})
  \quad \text{and} \quad
  \nabla \psi(\alpha_{t+1})
  =
  \nabla \psi(\alpha_{t})
  - \eta \nabla R_n(\alpha_{t}),
\end{equation}
where $\eta > 0$ is the step-size. We remark that the choice $\psi(\alpha) =
\frac{1}{2}\norm{\alpha}_{2}^{2}$ reduces the above updates to gradient
descent. A key notion in the analysis of mirror descent
algorithms is the \emph{Bregman divergence}, defined as
$
  D_{\psi}(\alpha', \alpha)
  =
  \psi(\alpha') - \psi(\alpha) - \ip{\nabla \psi(\alpha)}{\alpha' - \alpha}
$
for all $\alpha', \alpha$ in the domain of $\psi$.
By convexity of $\psi$, the Bregman divergence $D_{\psi}$ is non-negative and
enters the analysis of mirror descent algorithms through the
following elementary equality:
\begin{equation}
  \label{eq:change-in-potential-continuous}
  -\frac{d}{dt}D_{\psi}(\alpha', \alpha_{t})
  = \ip {-\nabla{R_n(\alpha_{t})}}{\alpha' - \alpha_{t}}.
\end{equation}
Let $\bar{\alpha}_{t} = \frac{1}{t}\myint{0}{t}{\alpha_{t}dt}$.
In the optimization literature, the above equation can be used to~establish
that $R_n(\bar{\alpha}_{t})$ can get arbitrarily close to
$R_n(\alpha')$ from above, for any reference point $\alpha'$.
In particular, by convexity of $R_n$,
we have $\ip {-\nabla{R_n(\alpha_{t})}}{\alpha' - \alpha_{t}}
\geq R_n(\alpha_{t}) - R_n(\alpha')$ and so
\begin{equation}
  \label{eq:potential-proof-one-line}
  \frac{1}{t}D_{\psi}(\alpha', \alpha_{0})
  \geq \frac{1}{t}\myint{0}{t}{-\frac{d}{ds}D_{\psi}(\alpha', \alpha_{s})ds}
  \geq \frac{1}{t}\myint{0}{t}{R_n(\alpha_{s}) - R_n(\alpha') ds}
  \geq
  R_n(\bar{\alpha}_{t}) - R_n(\alpha'),
\end{equation}
where the last line follows by convexity of $R_n$.
Remarkably, the above proof works independently of the choice of the mirror
map $\psi$, establishing convergence for a \emph{family} of algorithms in
a unified framework.
For more information we refer the interested reader to the surveys by
\citet{bubeck2015convex} and \citet{bansal2019potential}.
The latter survey focuses entirely on such potential-based
proofs in a variety of settings, including acceleration.

\section{Summary of Techniques and Main Results}
\label{section:our-work}

We develop a general theory for learning linear models (including kernel machines) with
the squared loss that shows how the optimization trajectory of
\emph{unconstrained} mirror descent applied to minimize the
unregularized empirical risk is \emph{inherently} connected to excess
risk guarantees via offset Rademacher complexity. Unlike in most prior work on
early stopping,
the notion of statistical complexity appears naturally from intrinsic
properties of mirror descent applied to the unregularized empirical risk,
without invoking lower-level arguments related to
concentration to the \emph{fictitious} population version of the algorithm.
Furthermore, our theory leads to an explicit characterization of stopping times
from the point of view of both optimization and statistics, which directly
yields excess risk bounds and allows us to re-derive previously established
results, and some new results, in a much simpler~fashion.

As discussed in Section~\ref{section:background},
early-stopped unconstrained iterative algorithms do not easily fit within the mathematical
framework of classical localization techniques, partially explaining the
scarcity of results connecting localized complexity measures with such algorithms.
Offset Rademacher complexities, on the other hand, open up another avenue for
establishing such connections via the design of update rules tailored to
satisfy the offset condition (cf.\ Definition~\ref{definition:offset-condition}).
Instead of optimizing the empirical risk $R_{n}$, a natural approach to
consider is an application of some iterative optimization algorithm to
directly minimize the term appearing in the definition of the offset condition:
$\tilde{R}^{{\alpha^\prime}\hspace{-0.2em},\hspace{0.15em}c}_{n}(\alpha) =R_{n}(\alpha) - R_{n}(\alpha') +
c\inlinenorm{g_{\alpha} - g_{\alpha'}}_{n}^{2}$.
For any $c > 0$, the gradient $\nabla_{\alpha} \tilde{R}^{{\alpha^\prime}\hspace{-0.2em},\hspace{0.15em}c}_n(\alpha)$
depends on the unknown reference point $\alpha'$ and hence cannot be
computed in practice.
Remarkably, we show that the mirror descent updates applied to the empirical loss
$R_{n}$ simultaneously \emph{implicitly} minimizes $\tilde{R}^{{\alpha^\prime}\hspace{-0.2em},\hspace{0.15em}1}_{n}$
for \emph{all} reference points $\alpha'$ up to a certain stopping time (which
depends on $\alpha'$), while also \emph{staying inside a certain Bregman
  ``ball''} centered at $\alpha'$ up to the corresponding stopping time. While mirror descent was developed within
the framework of convex optimization, it has also found applications
in a wide range of problems including
bandits \cite{abernethy2009competing},
online learning \cite{hazan2016introduction},
the k-server problem \cite{bubeck2018k}
and metrical task systems \cite{bubeck2019metrical}.
In this respect, our work can be seen as an exposition of yet another example
where mirror descent naturally solves a problem outside of its originally intended scope.

The key insight behind our main result is the following identity,
linking the potential-based analysis of mirror descent
(cf.\ Section~\ref{section:background})
to the statistical guarantees derived from offset complexities via the offset
condition (cf.\ Definition~\ref{definition:offset-condition}).
\begin{lemma}
  \label{lemma:missing-term}
  For any $\alpha, \alpha' \in \R^{m}$, the following holds:
  \[
    \ip{-\nabla R_{n}(\alpha)}{\alpha' - \alpha}
    =
    R_{n}(\alpha) - R_{n}(\alpha')
    +
    \inlinenorm{g_{\alpha} - g_{\alpha}'}_{n}^{2}.
  \]
\end{lemma}
\begin{proof}
  Recall that by the existence of some $Z \in \R^{n\times m}$ such that
  for any parameter $\alpha$ we have $g_{\alpha}(x_{i}) = (Z\alpha)_{i}$
  (cf.\ Section~\ref{section:introduction}),
  we can express the empirical loss function as
  $R_{n}(\alpha) = \frac{1}{n}\inlinenorm{Z\alpha - y}_{2}^{2}$,
  where $y \in \R^{n}$ is a vector with the $i$\th\ entry equal to $y_{i}$.
  Hence, for any $\alpha, \alpha' \in \R^{m}$ we have
  \begin{align*}
    \ip{-\nabla R_{n}(\alpha)}{\alpha' - \alpha}
    &=
    \frac{2}{n}\ip{-Z^{\mathsf{T}}(Z\alpha - y)}{\alpha' - \alpha} \\
    &=
    \frac{2}{n}\ip{-(Z\alpha -Z\alpha' + Z\alpha'- y)}{Z(\alpha' - \alpha)} \\
    &=
    \frac{2}{n}\norm{Z\alpha - Z\alpha'}_{2}^{2}
    - \frac{1}{n}\cdot 2
      \ip{Z\alpha'- y}{Z(\alpha' - \alpha)} \\
    &=
    \frac{2}{n}\norm{Z\alpha - Z\alpha'}_{2}^{2}
    - \frac{1}{n}\cdot\left(
        \norm{Z\alpha' - y}^{2}_{2}
      + \norm{Z(\alpha - \alpha')}_{2}^{2}
      - \norm{Z\alpha - y}^{2}_{2}
    \right) \\
    &=
    \frac{2}{n}\norm{Z\alpha - Z\alpha'}_{2}^{2}
    - \frac{1}{n}\cdot\left(
        nR_{n}(\alpha')
      + \norm{Z(\alpha - \alpha')}_{2}^{2}
      - nR_{n}(\alpha)
    \right) \\
    &= R_{n}(\alpha) - R_{n}(\alpha')
      + \norm{g_{\alpha} - g_{\alpha'}}_{n}^{2},
  \end{align*}
  where the fourth line follows by applying the equality
  $
    2\ip{a}{b} = \norm{a}^{2}_{2} + \norm{b}^{2}_{2} - \norm{a-b}^{2}_{2},
  $
  which holds for any vectors $a, b \in \R^{m}$.
\end{proof}
\begin{figure}[ht]
  \centering
  \begin{subfigure}{.45\textwidth}
    \centering
    \includegraphics[width=\linewidth]{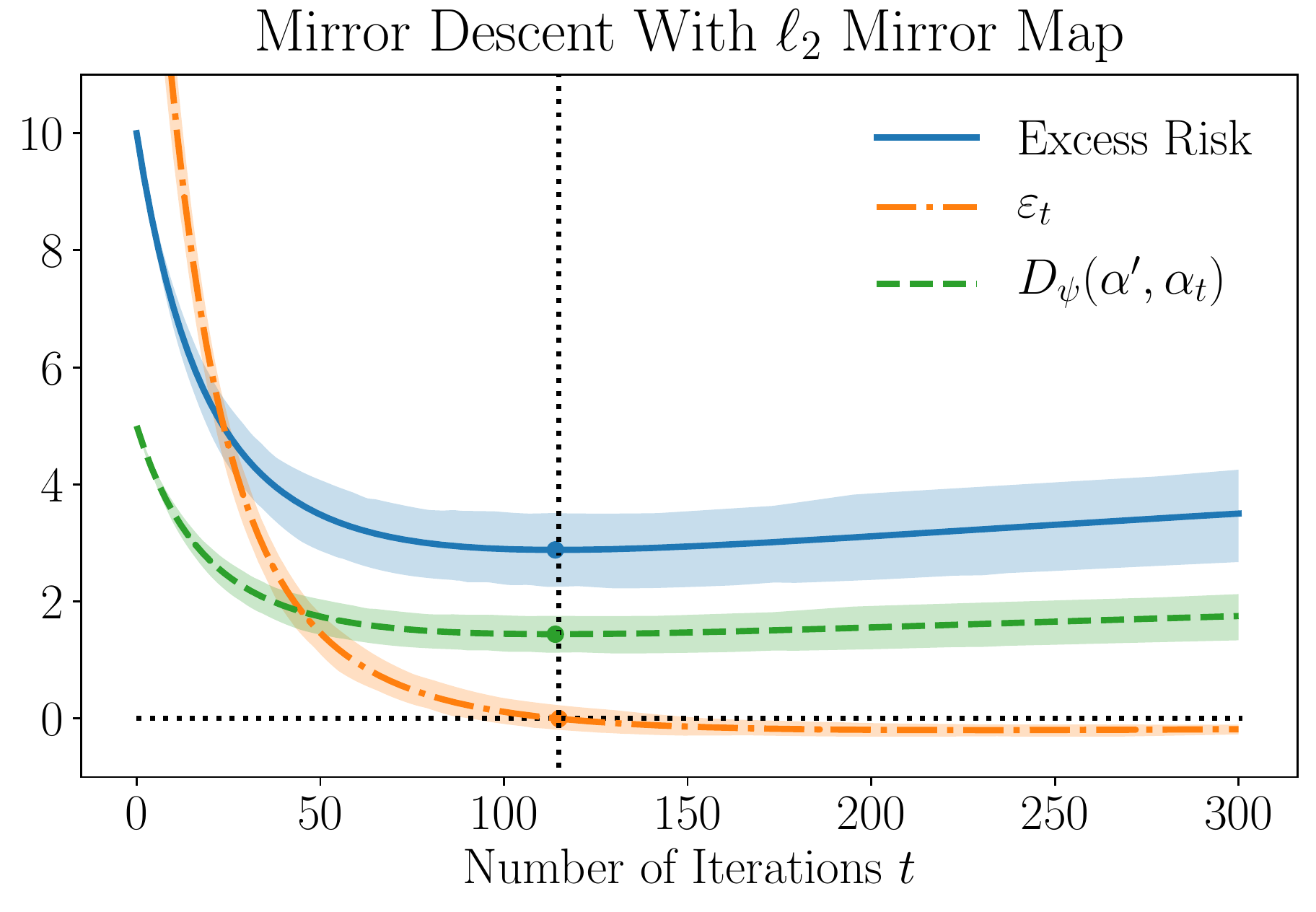}
  \end{subfigure}%
  \begin{subfigure}{.45\textwidth}
    \centering
    \includegraphics[width=\linewidth]{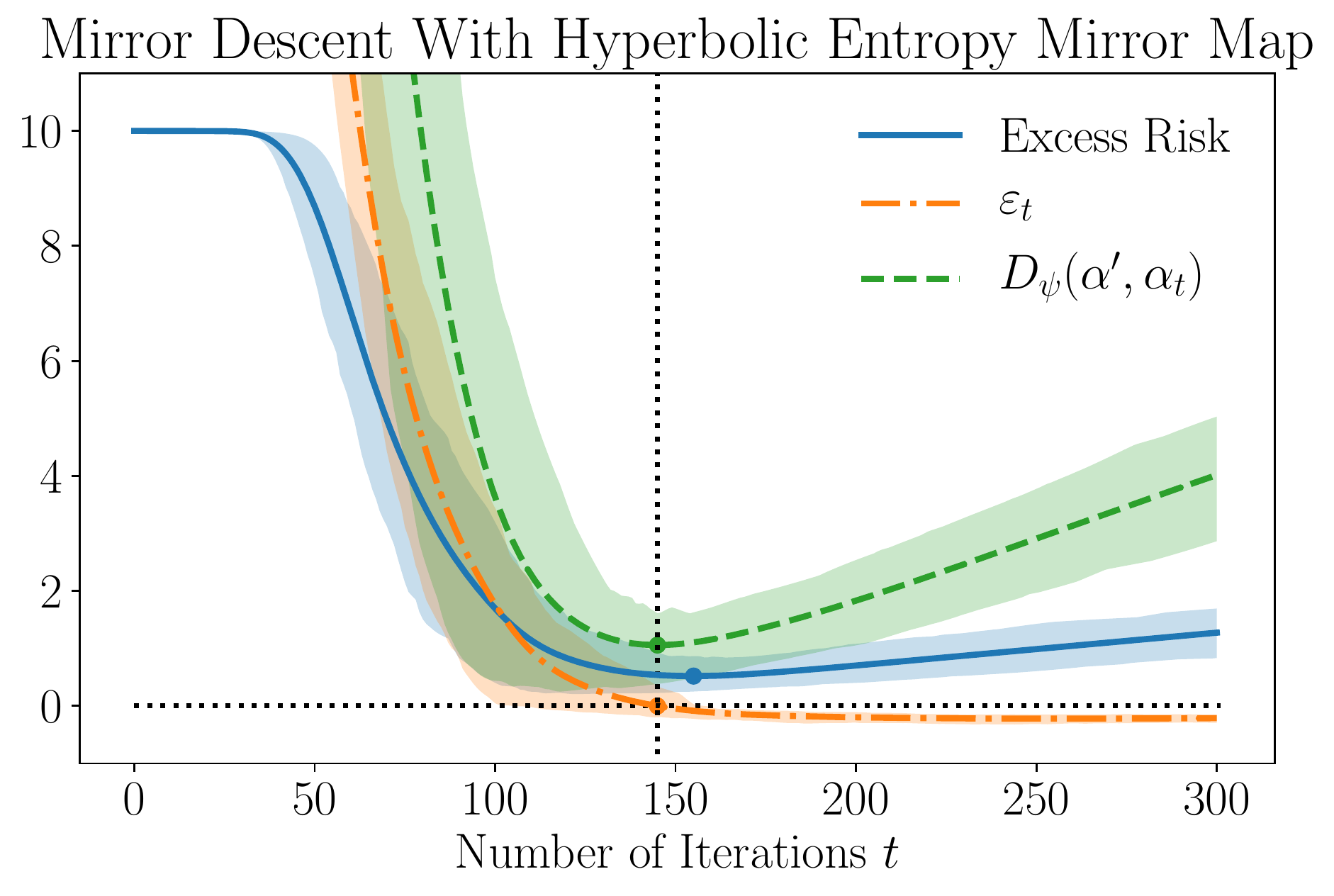}
  \end{subfigure}%
  \caption{Consider the setting of
    Figure~\ref{fig:explicit-vs-implicit-regularization} and let
    $\varepsilon_{t} = R_{n}(\alpha_{t}) - R_{n}(\alpha') +
    \norm{g_{\alpha_{t}} - g_{\alpha'}}_{n}^{2}$.
    The above plots illustrate the following two points.
    First, there exists a stopping time time $t^{\star}$
    such that $\varepsilon_{t^{\star}} \approx 0$ (denoted by the vertical
    dotted line). Hence, the triple $(P, \{g_{\alpha'}\}, g_{\alpha_{t^{\star}}})$
    satisfies the offset condition (cf.\ Definition~\ref{definition:offset-condition})
    with parameters $(c=1, \varepsilon \approx 0)$.
    Second, while $\varepsilon_{t} \geq 0$, the Bregman divergence
    $D_{\psi}(\alpha', \alpha_{t})$ denoted by the green line
    is non-increasing. It follows that the estimator
    $g_{\alpha_{t^{\star}}}$ is constrained to lie in the set
    $\{g_{\alpha} : D_{\psi}(\alpha', \alpha) \leq D_{\psi}(\alpha',
      \alpha_{0}) \}$, the offset complexity of which can be used to
      upper-bound the excess risk of interest.
    Crucially, this type of analysis does not directly rely on the particular
    form taken by the mirror descent update rules, which bypasses the
    limitations present in prior work (cf.\ Section~\ref{section:related-work})
    and allows us to
    provide excess risk guarantees for a family of mirror descent algorithms.
    In the plot above, the solid lines denote means over $100$ runs, the dots denote the
    minimum of each solid line, whereas the shaded regions
    correspond to the $10\th$ and the $90\th$ percentiles.}
  \label{fig:analysis-of-early-stopped-mirror-descent}
\end{figure}
To appreciate the significance of the above lemma we revisit the
potential-based proof of mirror descent presented in
Equation~\eqref{eq:potential-proof-one-line} in
Section~\ref{section:background}. This time, instead of using the
convexity of $R_{n}$ which gives
$\ip {-\nabla{R_n(\alpha_{t})}}{\alpha' - \alpha_{t}}
\geq R_n(\alpha_{t}) - R_n(\alpha')$, we directly plug in the
identity given in Lemma~\ref{lemma:missing-term} into
Equation~\eqref{eq:change-in-potential-continuous} which yields the following
\emph{equality}:
\begin{equation*}
  -\frac{d}{dt}D_{\psi}(\alpha', \alpha_{t})
  =
  R_{n}(\alpha_{t}) - R_{n}(\alpha') + \inlinenorm{g_{\alpha_{t}} - g_{\alpha'}}_{n}^{2}.
\end{equation*}
The above equation shows that while
$R_{n}(\alpha_{t}) - R_{n}(\alpha') + \inlinenorm{g_{\alpha_{t}} -
g_{\alpha'}}_{n}^{2} > 0$, the iterates of mirror descent stay withing the
Bregman ball $\{ \alpha \in \R^{m} : D_{\psi}(\alpha', \alpha) \leq
  D_{\psi}(\alpha', \alpha_{0})\}$. At the same time, the integration argument
used in Equation~\eqref{eq:potential-proof-one-line} establishes that
the term $R_{n}(\alpha_{t}) - R_{n}(\alpha') + \inlinenorm{g_{\alpha_{t}} -
g_{\alpha'}}_{n}^{2}$ eventually gets arbitrarily close to $0$, and thus
the early-stopped mirror descent iterates satisfy the offset
condition (cf.\ Definition~\ref{definition:offset-condition}).
For a visual demonstration of the above proof sketch see
Figure~\ref{fig:analysis-of-early-stopped-mirror-descent}.
We provide full details of this argument in the proof of
Theorem~\ref{thm:continuous-time} as well as a discrete-time version in
Theorem~\ref{thm:smooth-discretization}.

\vspace{9pt} % Right now the "Summary of contributions" is trailing at the
             % bottom of the page, hence the vspace to move it to the next
             % page.
\paragraph{Summary of contributions:}
\begin{enumerate}[topsep=0.5em,itemsep=0.1em,leftmargin=*]
  \item Our work extends the scope of offset Rademacher complexities to a
    family of early-stopped mirror descent methods. Additionally, we
    extend the scope of mirror descent to be used as a computationally efficient
    statistical device in an i.i.d.\ batch statistical learning setting.
	\item Our main results, in a short and transparent way, yield bounds on the
    excess risk of the iterates of (both continuous-time and discrete-time)
    mirror descent using offset Rademacher complexities. In contrast to
    prior work, our arguments require no direct use of low-level
    mathematical techniques such as symmetrization, peeling, or
    concentration to the population version of the algorithm.
	\item In Section~\ref{section:applications}, we demonstrate some selected
    applications of our main results and comment on the connections to the
    related work therein.
\end{enumerate}

\subsection{Comparison with Related Work}
\label{section:related-work}

Statistical and computational properties of unconstrained gradient descent
updates have been a subject of intense study over the past two decades,
with most of the existing results focusing on the quadratic loss in reproducing
kernel hilbert spaces (RKHS) \citep{
  buhlmann2003boosting,
  yao2007early,
  bauer2007regularization,
  raskutti2014early,
  wei2019early}.
In contrast to our work, the above work focuses either on bounds
in $\inlinenorm{\cdot}_{n}^{2}$ or in $\inlinenorm{\cdot}_{P}^{2}$ norms, which can be
arbitrarily smaller than the excess risk considered in our work
(see \cite[Section 1]{shamir2015sample} for an example).
In addition, the analysis in \citep{
  buhlmann2003boosting,
  yao2007early,
  bauer2007regularization,
  raskutti2014early}
is closely tied to the $\ell_{2}$ geometry of the gradient descent updates, which allows
one to view the algorithm as a particularly simple linear operator acting on the
observed labels. Spectral properties of these linear operators are then
analyzed as a function of the number of iterations, which can be solved
for a stopping time via some form of bias-variance decomposition.
Our work, in contrast, enables simultaneously studying a \emph{family} of
update rules, characterized by different choices of the mirror map,
in a unified framework without relying on their particular form.

One of the primary contributions of our work is the connection between
mirror descent iterates and localized complexity measures.
To the best of our knowledge, there are only two prior works making
connections of a similar nature, albeit only in the setting of Euclidean
gradient descent
updates, that is, with the choice of the mirror map $\psi(\alpha) =
\inlinenorm{\alpha}_{2}^{2}/2$ \citep{raskutti2014early, wei2019early}.
Such connections are observed in an algebraic fashion in the former work, while
localized complexities appear more naturally in \citep{wei2019early}, via
the analysis of the range of estimators defined by gradient descent iterates up
the stopping time. In this respect, the work in
\citep{wei2019early} is the closest to ours.
In Theorem~\ref{thm:nonparametric-regression}, we show how a straightforward
application of our main results immediately recovers results similar to the
ones obtained in \citep{raskutti2014early, wei2019early} and defer
an extended discussion of similarities and differences to
Section~\ref{section:application-nonparametric-regression}.

Beyond the Euclidean setup, interest in understanding the generalization
properties of neural networks has sparked research into \emph{implicit}
regularization properties of various factorized models.
In the context of neural networks, the authors of \citep{
  gunasekar2017implicit,
  li2018algorithmic,
  arora2019implicit,
  woodworth2019kernel,
  gidel2019implicit}
show that iterates of gradient descent applied to factorized matrix models
are implicitly biased towards some sparsity-inducing structure
such as low-rankness or low nuclear norm. Such results, however, hold
under certain limit statements, such as vanishing initialization or step-size,
the number of iterations going to infinity, or no noise in the problem.
In the setting of linear regression, matrix factorization models reduce to
vector Hadamard product factorizations, where
early-stopped gradient descent was shown to yield
minimax optimal rates for sparse recovery with the analysis vitally
relying on the restricted isometry property
\cite{zhao2019implicit, vaskevicius2019implicit}.
In Theorem~\ref{thm:l1-prediction-almost-optimal}, we demonstrate
a simple analysis of such updates within our framework
\emph{without any assumptions} on the design matrix other than bounded columns,
yielding
an (up to a log factor) minimax optimal algorithm for in-sample
linear prediction under $\ell_{1}$ norm~constraints.

Implicit regularization properties of mirror descent have recently attracted
a considerable amount of attention; however, most results in this area either focus on
optimization guarantees that do not provide any direct link to statistical
guarantees on out-of-sample prediction
\citep{gunasekar2018characterizing,azizan2018stochastic},
or establish a connection to statistics via some forms of explicit
regularization \citep{suggala2018connecting}.
The work \citep{suggala2018connecting} shows connections between the
iterates on the entire path and the
solutions on the regularization path for a suitable regularized risk
minimization problem. In Theorem~\ref{thm:implicit-vs-explicit},
we show how the analysis of such problems naturally fit into our framework
and defer an extended discussion and comparison to the other related work
\cite{ali2019continuous}
to Appendix~\ref{section:application-optimization-path}.
Yet other papers have used early stopping to solvers applied directly to appropriately
constrained problems and regularization-promoting structures encoded
directly into the loss function~\citep{matet2017don}.

Recent work has also focused on providing statistical guarantees for iterates
generated via gradient descent updates in
stochastic \citep{
  rosasco2015learning,
  lin2016generalization,
  neu2018iterate,
  ali2020implicit},
accelerated \citep{
  chen2018stability,
  pagliana2019implicit},
and distributed settings \citep{
  lin2018optimal,
  richards2019optimal,
  richards2020graph}.
These works provide statistical guarantees without establishing connections
to localized complexity measures;
we anticipate such connections to be studied within our framework in future
work, for a family of mirror descent algorithms.

\section{Main Results}
\label{section:main-results}
We first state and prove a continuous-time version of our main theorem,
which demonstrates the key ideas behind our approach in the simplest
setting. The first part of the theorem shows that the iterates of mirror
descent stay within a certain Bregman ball up to the prescribed stopping
time $t^{\star}$. The second part of the theorem immediately establishes
that when the parametrization given by $\alpha \in \R^{m}$ is independent of the data,
the early-stopped estimator $g_{\alpha_{t^{\star}}}$ satisfies the
offset condition (cf.\ Definition~\ref{definition:offset-condition}) with parameters
$c = 1$ and any $\varepsilon > 0$.\footnote{
  When the parametrization is data dependent, such as in the setting of
  kernel methods, our main theorems also establish that the early-stopped mirror descent
  iterates satisfy the offset condition (cf.\
  Definition~\ref{definition:offset-condition}).
  We analyze a concrete example and provide full details in
  Theorem~\ref{thm:nonparametric-regression}.
}
For the applications we consider, we choose $\varepsilon$ to match the complexity measure
of interest and recover the statistical-computational trade-offs consistent
with the previous results in the literature. In particular,
$t^{\star} = O(D_{\psi}(\alpha', \alpha_{0})/\varepsilon)$,
so that achieving higher statistical accuracy requires more computational
power. Finally, we note that the dependence of $t^{\star}$ on the unknown
radius $D_{\psi}(\alpha', \alpha_{0})$ is unavoidable purely from an optimization
point of view.
%------------------------------ Continuous Time -------------------------------
%------------------------------------------------------------------------------
\begin{theorem}
  \label{thm:continuous-time}
  Consider the continuous-time mirror descent updates given in
  Equation~\eqref{eq:mirror descent-updates}.
  Let $\alpha_{0}$ be the initialization point, $\alpha'$ be any chosen
  reference point, and fix any $\varepsilon > 0$.
  Then, there exists a stopping time
  $t^{\star} = t^{\star}(D_{n}, \psi, \alpha_{0}, \alpha') \leq
   2D_{\psi}(\alpha', \alpha_{0})/\varepsilon$ such that:
   \begin{enumerate}
    \item For all $0 \leq t \leq t^{\star}$,
      $g_{\alpha_{t}} \in
        \mathcal{G}(\psi, \alpha_{0}, \alpha')
        =
        \{ g_{\alpha} \in \R^{m} :
            D_{\psi}(\alpha', \alpha)
            \leq D_{\psi}(\alpha', \alpha_{0})
        \}.
      $
    \item At the stopping time $t^{\star}$, we have
      $R_{n}(\alpha_{t^{\star}}) - R_{n}(\alpha') +
      \inlinenorm{g_{\alpha_{t^{\star}}} - g_{\alpha'}}_{n}^{2} \leq \varepsilon$.
   \end{enumerate}
\end{theorem}
\begin{proof}
  To simplify the notation let
  $\delta_{t} =R_{n}(\alpha_{t}) - R_{n}(\alpha')$
  and $r_{t} =\inlinenorm{g_{\alpha_{t}} - g_{\alpha'}}_{n}^{2}$.
  Combining Equation~\eqref{eq:change-in-potential-continuous} with
  Lemma~\ref{lemma:missing-term} we have
  $-\frac{d}{dt}D_{\psi}(\alpha', \alpha_{t}) = r_{t} + \delta_{t}.$
  Let $T = 2D_{\psi}(\alpha', \alpha_{0}) / \varepsilon$.
  Integrating both sides of the above equality we obtain
  \begin{align*}
    &D_{\psi}(\alpha', \alpha_{0}) - D_{\psi}(\alpha', \alpha_{T})
    =
    \myint{0}{T}{-\frac{d}{dt}D_{\psi}(\alpha', \alpha_{t})dt}
    =
	 \myint{0}{T}{(r_{t} + \delta_{t})dt} \\
   \implies
    &
    \inf_{0 \leq t \leq T} \left\{ r_{t} + \delta_{t} \right\}
    \leq
    \frac{1}{T}\myint{0}{T}{
		 (r_{t} + \delta_{t}) dt
    }
    \leq
    \frac{D_{\psi}(\alpha', \alpha_{0})}{T}
    \leq \frac{\varepsilon}{2}.
  \end{align*}
  It follows that the following infimum is well defined:
  $
    t^{\star} = \inf\{0 \leq t \leq T \mid
      r_{t} + \delta_{t}
      \leq \varepsilon
    \}.
  $
  Hence, $r_{t^{\star}} + \delta_{t^{\star}} \leq \varepsilon$
  and for all $0 \leq t \leq t^{\star}$ we have
  \begin{align*}
    D_{\psi}(\alpha', \alpha_{0}) - D_{\psi}(\alpha', \alpha_{t})
    =
	 \myint{0}{t}{(r_{t} + \delta_{t}) dt}
    \geq t\varepsilon \geq 0.
  \end{align*}
  The above inequality implies that
  $D_{\psi}(\alpha', \alpha_{t}) \leq D_{\psi}(\alpha', \alpha_{0})$,
  which concludes our proof.
\end{proof}
%------------------------------- Discrete Time --------------------------------
%------------------------------------------------------------------------------
In the next theorem, we prove an equivalent result in discrete-time.
Let $\inlinenorm{\cdot}$ denote any norm.
We say that $R_{n}$ is $\beta$-smooth with respect to $\norm{\cdot}$ if
$R_{n}(\alpha') \leq R_{n}(\alpha) + \inlineip{\nabla R_{n}(\alpha)}{\alpha' -
  \alpha} + \frac{\beta}{2} \norm{\alpha - \alpha'}^{2}$
for any $\alpha, \alpha'$ in the domain of $R_{n}$.
We also say that the mirror map $\psi$ is $\rho$-strongly convex
with respect to $\inlinenorm{\cdot}$ if for any $\alpha, \alpha'$ we have
$ D_{\psi}(\alpha', \alpha) \geq \frac{\rho}{2}\norm{\alpha' - \alpha}^{2}$.
%--------------------------- Smooth Discretization ----------------------------
%------------------------------------------------------------------------------
\begin{theorem}
  \label{thm:smooth-discretization}
  Consider the discrete-time mirror descent updates given in
  Equation~\eqref{eq:mirror descent-updates}.
  Suppose that $R_{n}$ is $\beta$-smooth and $\psi$ is $\rho$-strongly
  convex with respect to some norm $\norm{\cdot}$.
  Let $\alpha_{0}$ be the initialization point, $\alpha'$ be any
  reference point, $\eta \leq \rho/\beta$, and fix any $\varepsilon > 0$.
  Then, there exists a stopping time
  $t^{\star} = t^{\star}(D_{n}, \psi, \alpha_{0}, \alpha', \eta) \leq
  (D_{\psi}(\alpha', \alpha_{0}) + \eta R_{n}(\alpha'))/(\eta\varepsilon)$
   such~that:
   \begin{enumerate}
    \item For all $0 \leq t \leq t^{\star}$,
      $g_{\alpha_{t}} \in
        \mathcal{G}(\psi, \alpha_{0}, \alpha', \eta)
        =\{ g_{\alpha} :
            D_{\psi}(\alpha', \alpha)
            \leq D_{\psi}(\alpha', \alpha_{0}) + \eta R_{n}(\alpha')
          \}$.

    \item At the stopping time $t^{\star}$, we have
      $R_{n}(\alpha_{t^{\star}}) - R_{n}(\alpha') +
      \inlinenorm{g_{\alpha_{t^{\star}}} - g_{\alpha'}}_{n}^{2} \leq \varepsilon$.
   \end{enumerate}
\end{theorem}
Before providing the proof we briefly comment on the above theorem. First, the
step-size
condition $\eta \leq \rho/\beta$ and the number of iterations
$O(1/\varepsilon)$ needed to reach a desired level of accuracy
are identical to the guarantees proved in purely convex optimization settings
(cf.\ Theorem 4.4 in \cite{bubeck2015convex}).
On the other hand, comparing Theorems~\ref{thm:continuous-time} and
\ref{thm:smooth-discretization}, in the discrete setting we pay a
price of $\eta R_{n}(\alpha')$ in the radius of the Bregman ball
where our early-stopped estimator lies.
This is consistent with prior work in the early stopping literature,
where such an expansion of the radius dependent on the noise level\footnote{
  Since $\alpha'$ is independent of the data and since it
  corresponds to the best parameter in some class of interest,
  $R_{n}(\alpha') \approx R(\alpha')$ and hence it can be interpreted as the
  noise level of the problem.
} propagates
into the resulting bounds (cf.\ definition of $C$ in Theorem 1 in
\cite{wei2019early}). Our work, on the other hand, allows for a
more fine-grained control of statistical-computational trade-offs
via a selection of a small enough step-size $\eta$.

We now introduce two lemmas supporting the proof of
Theorem~\ref{thm:smooth-discretization}.
The first lemma is a well-known generalization of the
Euclidean identity
$\inlinenorm{a}_{2}^{2} + \inlinenorm{b}_{2}^{2} =
 \inlinenorm{a-b}_{2}^{2} + 2\inlineip{a}{b}$, which holds for
any Bregman divergence $D_{\psi}$ induced by any mirror map $\psi$.

\begin{lemma}
  \label{lemma:law-of-cosines-for-bregman-divergences}
  For any mirror map $\psi$ and any points $x,y,z$ in the domain of $\psi$
  we have
  $$
    D_{\psi}(z, x) - D_{\psi}(z, y)
    =
    \ip{\nabla \psi(x) - \nabla \psi(y)}{x - z}
    - D_{\psi}(x, y).
  $$
\end{lemma}
\begin{proof}
  The identity follows by the definition of Bregman divergence.
\end{proof}

The second lemma proves a discrete-time counterpart to the
identity given in Equation~\eqref{eq:change-in-potential-continuous},
which combined with Lemma~\ref{lemma:missing-term} states
that $-\frac{d}{dt}D_{\psi}(\alpha', \alpha_{t})
= \ip {-\nabla{R_n(\alpha_{t})}}{\alpha' - \alpha_{t}} = \delta_{t} + r_{t}$.
We remind our reader that $\delta_{t} = R_{n}(\alpha_{t}) - R_{n}(\alpha')$
and $r_{t} = \inlinenorm{g_{\alpha_{t}} - g_{\alpha'}}_{n}^{2}$.
\begin{lemma}
  \label{lemma:discretized-change-in-potential}
  Consider the discrete-time mirror descent updates given in
  Equation~\eqref{eq:mirror descent-updates}.
  Suppose that $R_{n}$ is $\beta$-smooth and $\psi$ is $\rho$-strongly
  convex with respect to some norm $\inlinenorm{\cdot}$,
  and the step-size $\eta$ satisfies $\eta \leq \rho/\beta$.
  Then, the following inequality holds for all $t \geq 0$:
  \begin{equation*}
    D_{\psi}(\alpha', \alpha_{t}) - D_{\psi}(\alpha', \alpha_{t+1})
    \geq \eta(\delta_{t+1} + r_{t}).
  \end{equation*}
\end{lemma}

\begin{proof}
  Combining Lemma~\ref{lemma:law-of-cosines-for-bregman-divergences} with
  the definition of discrete-time mirror descent updates (cf.\
  Equation~\eqref{eq:mirror descent-updates}) we have
  \begin{align}
    &\phantom{{}={}}D_{\psi}(\alpha', \alpha_{t}) - D_{\psi}(\alpha', \alpha_{t+1}) \notag \\
    &=
    \ip{\nabla \psi(\alpha_{t}) - \nabla \psi(\alpha_{t+1})}{\alpha_{t} - \alpha'}
    - D_{\psi}(\alpha_{t}, \alpha_{t+1}) \notag \\
    &=
    \ip{\eta \nabla R_{n}(\alpha_{t})}{\alpha_{t} - \alpha'}
    - \left(
       \psi(\alpha_{t}) - \psi(\alpha_{t+1})
       - \ip{\nabla \psi(\alpha_{t+1})}{\alpha_{t} - \alpha_{t+1}}
      \right) \notag \\
    &=
    \ip{\eta \nabla R_{n}(\alpha_{t})}{\alpha_{t} - \alpha'}
    - \left(-D_{\psi}(\alpha_{t+1}, \alpha_{t})
      + \ip{\nabla \psi(\alpha_{t}) - \nabla \psi(\alpha_{t+1})}
           {\alpha_{t} - \alpha_{t+1}}
      \right) \notag \\
    &=
    \ip{\eta \nabla R_{n}(\alpha_{t})}{\alpha_{t} - \alpha'}
    - \left(-D_{\psi}(\alpha_{t+1}, \alpha_{t})
      + \ip{\eta\nabla R_{n}(\alpha_{t})}
           {\alpha_{t} - \alpha_{t+1}}
      \right) \notag \\
    &=
    \ip{\eta \nabla R_{n}(\alpha_{t})}{\alpha_{t} - \alpha'}
    + D_{\psi}(\alpha_{t+1}, \alpha_{t})
    + \ip{-\eta \nabla R_{n}(\alpha_{t})}{\alpha_{t} - \alpha_{t+1}}.
    \label{eq:discrete-change-in-potential-proof-equation}
  \end{align}
  By the $\rho$-strong convexity of the mirror map $\psi$, the second term
  in Equation~\eqref{eq:discrete-change-in-potential-proof-equation} can be lower-bounded as
  $D_{\psi}(\alpha_{t+1}, \alpha_{t}) \geq \frac{\rho}{2}\norm{\alpha_{t+1} -
    \alpha_{t}}^{2}$.
  The last term in
  Equation~\eqref{eq:discrete-change-in-potential-proof-equation}
  can be lower-bounded by using the $\beta$-smoothness condition of the
  empirical risk function $R_{n}$, which yields
  $
    \inlineip{-\nabla R_{n}(\alpha_{t})}{\alpha_{t} - \alpha_{t+1}}
    = \inlineip{\nabla R_{n}(\alpha_{t})}{\alpha_{t+1} - \alpha_{t}}
    \geq R_{n}(\alpha_{t+1}) - R_{n}(\alpha_{t}) - \frac{\beta}{2}\inlinenorm{
      \alpha_{t+1} - \alpha_{t}}^{2}.
  $
  We can hence continue from
  Equation~\eqref{eq:discrete-change-in-potential-proof-equation} as follows:
  \begin{align*}
    &\phantom{{}={}}D_{\psi}(\alpha', \alpha_{t}) - D_{\psi}(\alpha', \alpha_{t+1}) \\
    &=
    \ip{\eta \nabla R_{n}(\alpha_{t})}{\alpha_{t} - \alpha'}
    + D_{\psi}(\alpha_{t+1}, \alpha_{t})
    + \ip{-\eta \nabla R_{n}(\alpha_{t})}{\alpha_{t} - \alpha_{t+1}} \\
    &\geq
    \ip{\eta \nabla R_{n}(\alpha_{t})}{\alpha_{t} - \alpha'}
    + \frac{\rho}{2}\norm{\alpha_{t+1} - \alpha{t}}^{2}
    + \eta \left(
      R_{n}(\alpha_{t+1}) - R_{n}(\alpha_{t}) - \frac{\beta}{2}\inlinenorm{
      \alpha_{t+1} - \alpha_{t}}^{2}.
    \right) \\
    &=
    \ip{\eta \nabla R_{n}(\alpha_{t})}{\alpha_{t} - \alpha'}
    + \left(\frac{\rho - \eta\beta}{2}\right)\norm{\alpha_{t+1} - \alpha{t}}^{2}
    + \eta (\delta_{t+1} - \delta_{t}).
  \end{align*}
  Since $\eta \leq \rho/\beta$, the second term is lower-bounded by $0$.
  Also, by Lemma~\ref{lemma:missing-term}, the first term becomes
  $\eta\inlineip{-\nabla R_{n}(\alpha_{t})}{\alpha' - \alpha_{t}}
   = \eta(\delta_{t} + r_{t})$.
  Combining these two observations with the last equation above we obtain
  $$
    D_{\psi}(\alpha', \alpha_{t}) - D_{\psi}(\alpha', \alpha_{t+1})
    \geq
    \eta(\delta_{t+1} + r_{t}),
  $$
  which completes our proof.
\end{proof}

With Lemma~\ref{lemma:discretized-change-in-potential} at hand,
we can prove Theorem~\ref{thm:smooth-discretization} following along the
same steps used to prove Theorem~\ref{thm:continuous-time}, albeit with
the continuous-time equation
$-\frac{d}{dt}D_{\psi}(\alpha', \alpha_{t}) = \delta_{t} + r_{t}$
replaced with its discrete-time counterpart
$
    D_{\psi}(\alpha', \alpha_{t}) - D_{\psi}(\alpha', \alpha_{t+1})
    \geq
    \eta(\delta_{t+1} + r_{t}).
$
In the discrete-time equation, $\delta_{t}$ is replaced with
$\delta_{t+1}$, which results in the expansion of the radius of the
Bregman ball in which the mirror descent iterates lie before the prescribed
stopping time (cf.\ the discussion following the statement of
Theorem~\ref{thm:smooth-discretization} above).

\begin{proof}[Proof of Theorem~\ref{thm:smooth-discretization}]
  By Lemma~\ref{lemma:discretized-change-in-potential} we have
  $D_{\psi}(\alpha', \alpha_{t}) - D_{\psi}(\alpha', \alpha_{t+1})
    \geq
    \eta(\delta_{t+1} + r_{t})$.
  Let $
  T = \left\lceil
    \frac{D_{\psi}(\alpha', \alpha_{0}) + \eta R_{n}(\alpha')}
    {\eta \varepsilon}
    \right\rceil$.
  Summing both sides of the above equation for
  $t=0, \dots, T$ we obtain
  \begin{align*}
    &D_{\psi}(\alpha', \alpha_{0}) - D_{\psi}(\alpha', \alpha_{T})
    \geq
     \eta r_{0}
    + \sum_{t=1}^{T} \eta(r_{t} + \delta_{t})
    + \eta \delta_{T+1} \\
    \implies
    &
    \min_{t = 1, \dots, T} \left\{\delta_{t} + r_{t}\right\}
    \leq \frac{\sum_{t=1}^{T} r_{t} + \delta_{t}}{T}
    \leq \frac{D_{\psi}(\alpha', \alpha_{0}) + \eta R_{n}(\alpha')}{\eta T}
    \leq \varepsilon,
  \end{align*}
  where in the last line we have used the definition of $T$ and facts
  that $D_{\psi}(\alpha', \alpha_{T}) \geq 0$, $r_{0} \geq 0$,
  and $\delta_{T+1} \geq - R_{n}(\alpha')$.

  It follows that the following minimum is well defined:
  $t^{\star} = \min \{t = 0, \dots, T \mid r_{t} + \delta_{t} \leq \varepsilon\}.$
  Hence, $r_{t^{\star}} + \delta_{t^{\star}} \leq \varepsilon$,
  which completes the second part of the theorem.
  To complete the first part of the theorem, note that
  for any $1 \leq t \leq t^{\star}$ by
  telescoping the equation $
    D_{\psi}(\alpha', \alpha_{t}) - D_{\psi}(\alpha', \alpha_{t+1})
    \geq
    \eta(\delta_{t+1} + r_{t})
  $
  from $0$ to $t-1$ we obtain
  \begin{align*}
    &D_{\psi}(\alpha', \alpha_{0}) - D_{\psi}(\alpha', \alpha_{t})
    \geq
     \eta r_{0}
    + \sum_{t=1}^{t-1} \eta(r_{t} + \delta_{t})
    + \eta \delta_{t} \\
    \implies&
    D_{\psi}(\alpha', \alpha_{t})
    \leq
    D_{\psi}(\alpha', \alpha_{0})
    - \sum_{t=1}^{t-1} \eta(r_{t} + \delta_{t})
    - \eta \delta_{t}
    \leq
    D_{\psi}(\alpha', \alpha_{0}) + \eta R_{n}(\alpha'),
  \end{align*}
  where in the last line we have used the facts that $\delta_{t} + r_{t} > \varepsilon > 0$
  and $-\delta_{t} \leq R_{n}(\alpha')$.
\end{proof}

\section{Selected Applications of the Main Results}
\label{section:applications}

In this section, we discuss three selected applications of our main
theorems, each of which is meant to demonstrate a particular facet of our
framework as we discuss below.

Most of the results on early stopping in prior literature are
shown in the non-parametric regression setting over RKHS
(cf.\ Section~\ref{section:related-work}).
In such settings, the parametrization $\alpha$ depends on the observed
data. Theorem~\ref{thm:nonparametric-regression} that we present in
Section~\ref{section:application-nonparametric-regression}
demonstrates that such data-dependent parametrizations easily fit within our framework.
Additionally, we obtain results that in some ways improve upon related
work, e.g., we obtain bounds on excess risk with no assumptions on the
distribution $P$ other than boundedness.

In Section~\ref{section:application-l1}
we prove Theorem~\ref{thm:l1-prediction-almost-optimal} to demonstrate
two features of our main results. First, via a proper choice of the mirror map,
we can derive a nearly minimax optimal algorithm for linear prediction under $\ell_{1}$
norm constraints. Our choice of the mirror map models the updates previously studied
in the early stopping literature in the works \cite{zhao2019implicit,
  vaskevicius2019implicit}, both of which heavily rely on the restricted
  isometry assumption. In contrast, our work only requires that the columns
  of the fixed-design matrix are bounded in $\ell_{2}$ norm. Hence,
Theorem~\ref{thm:l1-prediction-almost-optimal}
expands the scope of the results proved in \cite{zhao2019implicit, vaskevicius2019implicit}.
Second, we demonstrate that not only is our framework suitable to study
excess risk bounds, but our main results can also be flexibly adapted to
the setting of interest, in this case, proving bounds on the~in-sample~prediction~error.

Some recent work \citep{suggala2018connecting, ali2019continuous}
investigated the connections between \emph{continuous-time}
optimization paths traced by gradient and mirror descent algorithms, and
regularization paths of suitably regularized problems.
Via Theorem~\ref{thm:implicit-vs-explicit} proved in
Section~\ref{section:application-optimization-path}, we demonstrate that
such questions can also be addressed within our framework.
In contrast, the prior work drawing connections between early stopping
and \emph{classical} localized complexity measures \cite{raskutti2014early,
wei2019early} is unlikely to be easily adjustable to approach such questions
due to the limitations that we have outlined in Section~\ref{section:background}.

Crucially, various different questions recently studied in the related
literature naturally fit within the framework developed in our paper,
which provides a simple and unified way to approach such problems.
Moreover, in all of the three considered examples,
we prove results that in some aspects improve upon the prior work spanning
several different sub-areas in the early stopping literature.

\subsection{Early Stopping for Non-Parametric Regression}
\label{section:application-nonparametric-regression}

Let $P$ be \emph{any} distribution supported on $\mathcal{X} \times [-M, M]$
and let $k : \mathcal{X} \times \mathcal{X} \to [0, \infty)$
be a Mercer kernel which induces a Hilbert space of functions $\mathcal{H}$
equipped with norm $\inlinenorm{\cdot}_{\mathcal{H}}$.
Assume that $\sup_{x \in \mathcal{X}} k(x, x) \leq L$ for some constant $L > 0$
and, conditionally on the observed
data, denote by $K \in \R^{n \times n}$ a matrix such that
$K_{ij} = k(x_{i}, x_{j})$.
Such a setup is standard in the literature and we refer the interested reader
to the book by \citet{scholkopf2001learning} for more background on RKHS.

In the theorem below, we consider the updates defined as
\begin{equation}
  \label{eq:kernel-updates}
  \alpha_{0} = 0,\qquad
  \alpha_{t+1} = \alpha_{t} - \frac{\eta}{n}(K\alpha_{t} - y),
\end{equation}
with $\eta \leq 1 \wedge 1/\lambda_{max}(K/n)$, where
$\lambda_{\max}(K/n)$ is the maximum eigenvalue of $K/n$.
We let $Z = K$ so that $R_{n}(\alpha) = \inlinenorm{Z\alpha -
y}_{2}^{2}/n$ and thus the updates given in \eqref{eq:kernel-updates}
correspond to mirror descent updates with the mirror map
$\psi(\alpha) = \alpha^{\mathsf{T}}K\alpha$.\footnote{
  Typically, the map $\nabla \psi : \R^{n} \to \R^{n}$ is required to be
  surjective to ensure that the elements of the dual space can always be
  mapped back to the primal space, which is not necessarily the case with
  the choice of the mirror map $\psi(\alpha) = \alpha^{\mathsf{T}}K\alpha$.
  However, note that pre-multiplying both sides of
  Equation~\eqref{eq:kernel-updates} by $2K$, for any $t \geq 0$ it holds that
  $\nabla \psi(\alpha_{t+1}) = \nabla \psi(\alpha_t) - \eta \nabla
  R_{n}(\alpha_{t})$ and hence the updates defined in \eqref{eq:kernel-updates}
  are mirror descent updates.
} Note that $R_{n}$ is $2\lambda_{\text{max}}(K/n)$-smooth with respect to
the $\inlinenorm{\cdot}_{K}$ norm defined as $\norm{\alpha}_{K}^{2} =
\alpha^{\mathsf{T}}K\alpha$. To see that, note that any $\alpha, \alpha'$
by Lemma~\ref{lemma:missing-term} we have
\begin{align*}
  R_{n}(\alpha') - R_{n}(\alpha) - \ip{\nabla R_{n}(\alpha)}{\alpha' - \alpha}
  &= \frac{1}{n}\inlinenorm{K(\alpha - \alpha')}_{2}^{2}
  \leq \frac{1}{n}\inlinenorm{\sqrt{K}}^{2}\inlinenorm{\sqrt{K}(\alpha' - \alpha)}_{2}^{2}
  \\
  &= \frac{1}{n}\lambda_{\max}(K)\norm{\alpha' - \alpha}_{K}^{2}.
\end{align*}
Also, the mirror map
  $\psi(\alpha) = \inlinenorm{\alpha}_{K}^{2}$ is $2$-strongly convex
with respect to the $\inlinenorm{\cdot}_{K}$ norm.
Hence the choice of the step-size given above satisfies
the pre-conditions of Theorem~\ref{thm:smooth-discretization}.

To each $\alpha \in \R^{n}$, we associate a $g_{\alpha} \in \mathcal{H}$
defined as
$g_{\alpha} = \sum_{i=1}^{n} \alpha_{i}k(x_{i}, \cdot)$.
For any $\alpha, \alpha'$, the squared distance between the
functions $g_{\alpha'}$ and $g_{\alpha}$ with respect to the
RKHS norm $\inlinenorm{\cdot}_{\mathcal{H}}$ is given by
the Bregman divergence $D_{\psi}(\alpha', \alpha)$:
$$
  \inlinenorm{g_{\alpha'} - g_{\alpha}}_{\mathcal{H}}^{2}
  = (\alpha' - \alpha)^{\mathsf{T}}K(\alpha' - \alpha)
  = \inlinenorm{\alpha' - \alpha}_{K}^{2}
  = D_{\psi}(\alpha', \alpha).
$$

Since our parameter system is data-dependent (both $K$ and the parametrization given by
$\alpha$ depend on the observed data points), there is, in general, no single
$\alpha' \in \R^{n}$ such that $g' = g_{\alpha'}$
for all realization of $D_{n}$, where $g' \in \mathcal{H}$ is some arbitrary
reference function of interest. Hence, Theorem~\ref{thm:smooth-discretization}
does not immediately establish that early-stopped mirror descent iterates
satisfy the offset condition. Our proof that we present below
demonstrates how a data-dependent parameter system can be analyzed within our framework.
The key idea is to find $\alpha'(D_{n}) \in \R^{n}$, one for each dataset $D_{n}$,
such that $g_{\alpha'(D_{n})}$ is ``close enough'' to a reference function~of~interest~$g'$.

\begin{theorem}
  \label{thm:nonparametric-regression}
  Consider the setup described above. Fix any $R > 0$
  and let $\mathcal{F}_{R} = \{ h \in \mathcal{H} : \norm{h}_{\mathcal{H}} \leq R\}$.
  There exists a data-dependent stopping time $t^{\star} \leq
  c_{3}/(\eta \ex{\mathfrak{R}_{n}(\mathcal{F}_{R}, c_{2})})$
  such that
  $$
    \ex{\mathcal{E}(g_{\alpha_{t^{\star}}}, \mathcal{F}_{R})}
    \leq
    c_{1}\ex{\mathfrak{R}_{n}(\mathcal{F}_{R}, c_{2})},
  $$
  where constants $c_{1}, c_{2}, c_{3} > 0$ depend only on
  the boundedness constants $M, L$ and $R$.
\end{theorem}

Before presenting the proof, we compare the above theorem
with the related work connecting early stopping and localized complexity measures
\citep{raskutti2014early, wei2019early} in the setting of RKHS.
First, the bounds obtained in \citep{raskutti2014early, wei2019early} hold with
high-probability rather than in expectation, as we prove in
Theorem~\ref{thm:nonparametric-regression}. We note that high-probability
results can also be readily obtained in our setting, via an application
of \citep[Theorem 4]{liang2015learning},
for classes that satisfy the lower-isometry assumption
(cf.\ \citep[Definition 1]{liang2015learning}), which in some cases can hold
even in heavy-tailed setups. Thus, while there is some loss in our results
being stated in expectation, we can also obtain high-probability results that
sometimes fall outside of the scope of the sub-Gaussian setting considered in
\citep{raskutti2014early, wei2019early}.
Second, the work of \cite{wei2019early} considers a more general class of
loss functions, characterized by certain strong-convexity and smoothness
assumptions imposed at the population level, the reason being that the
classical theory of localized complexities allows for a fairly general class of
loss functions. On the other hand, while the theory of offset complexities
has been developed only for the quadratic loss thus far,
it allows us to consider \emph{a general class of algorithms}, as characterized by the
mirror map, and, in contrast to the works of \citep{raskutti2014early,
  wei2019early}, our main results can also be applied to provide statistical
guarantees \emph{along the whole optimization path} (cf.\
Theorem~\ref{thm:implicit-vs-explicit}).
Third, the bounds in \citep{raskutti2014early, wei2019early} were proved
under a well-specified model and i.i.d.\ noise assumptions, neither of which
is present in Theorem~\ref{thm:nonparametric-regression} considered in this
section. Finally, we provide bounds on the excess risk,
in contrast to bounds in $\inlinenorm{\cdot}_{P}^{2}$ norm considered in
\citep{raskutti2014early, wei2019early}, which coincide with the excess
risk only in some limited settings (cf.\ discussion in \citep{shamir2015sample}).
The differences aside, Theorem~\ref{thm:nonparametric-regression} is remarkably
similar to the results obtained in \citep{raskutti2014early, wei2019early}.
In particular, we recover similar conditions on the step-size and provide
almost identical statistical and computational guarantees.
We refer to \citep[Corollary 12]{liang2015learning} and \citep[Section 3.3]{wei2019early}
for discussions concerning the statistical optimality of localized complexity measures for
non-parametric regression.

%----------------------------------- Proof ------------------------------------
%------------------------------------------------------------------------------
\begin{proof}[Proof of Theorem~\ref{thm:nonparametric-regression}]
  By the Representer theorem, there exists $\alpha' = \alpha'(D_{n})$, such
  that $g_{\alpha'(D_{n})} \in
      \operatorname{argmin}_{g \in \mathcal{F}_{R}}\, R_{n}(g)$
  and hence, by convexity of $\mathcal{F}_{R}$, the triple $(P, \mathcal{F}_{R},
  g_{\alpha'(D_{n})})$ satisfies the offset condition with parameters
  ($c = 1$, $\varepsilon = 0$) (cf.\ \citep[Lemma 1]{liang2015learning}).
  Consequently, with probability $1$ we~have
  \begin{equation}
    \label{eq:erm-kernel-offset}
    R_{n}(\alpha'(D_{n})) - R_{n}(g_{\mathcal{F}_{R}})
    + \inlinenorm{g_{\alpha'(D_{n})} - g_{\mathcal{F}_{R}}}_{n}^{2}
    \leq 0.
  \end{equation}
  Since $\alpha_{0} = 0$, we have
  $D_{\psi}(\alpha'(D_{n}), \alpha_{0}) =
    \inlinenorm{\alpha'(D_{n})}_{K}^{2} =
    \inlinenorm{g_{\alpha'(D_{n})}}_{\mathcal{H}}^{2}
    \leq R^{2}
  $.
  Also, by the boundedness assumption on the kernel $k$, for any $g \in \mathcal{F}_{R}$ we have
  $\sup_{x} \abs{g(x)} \leq \inlinenorm{g}_{\mathcal{H}}L \leq RL$. Hence,
  for any $g \in \mathcal{F}_{R}$ we have
  $R_{n}(g) \leq (RL + M)^{2}$ and in particular,
  $R_{n}(\alpha'(D_{n})) \leq (RL + M)^{2}.$\footnote{
    In a well-specified setting this quantity can be made as small as the
    variance of the noise modulo an absolute multiplicative constant.
  }
  Applying Theorem~\ref{thm:smooth-discretization} with
  $\alpha' = \alpha'(D_{n})$, for any $\varepsilon > 0$, there exists
  a data-dependent stopping time $t^{\star}$ satisfying
  $$
    t^{\star} \leq (D_{\psi}(\alpha', \alpha_{0}) + \eta
    R_{n}(\alpha'))/(\eta \varepsilon)
         \leq (R^{2} + \eta(LR + M)^{2})/(\eta\varepsilon)
         \leq (R^{2} + (LR + M)^{2})/(\eta\varepsilon),
  $$
  such that the following two inequalities hold with probability $1$:
  \begin{align}
    &R_{n}(\alpha_{t^{\star}}) - R_{n}(\alpha'(D_{n}))
    + \inlinenorm{g_{\alpha_{t^{\star}}} - g_{\alpha'(D_{n})}}_{n}^{2}
    \leq \varepsilon, \notag \\
    &D_{\psi}(\alpha'(D_{n}), \alpha_{t^{\star}})
    \leq
    D_{\psi}(\alpha'(D_{n}), \alpha_{0}) + (LR + M)^{2},
    \label{eq:rkhs-norm-bound-bregman-divergence}
  \end{align}
  where the second inequality follows from the fact that
  $\eta R_{n}(\alpha'(D_{n})) \leq (LR + M)^{2}$.
  Combining the first inequality above with
  Equation~\eqref{eq:erm-kernel-offset},
  the following holds with~probability~$1$:
  \begin{align*}
    &R_{n}(\alpha_{t^{\star}}) - R_{n}(g_{\mathcal{F}_{R}})
    + \frac{1}{2}\inlinenorm{g_{\alpha_{t^{\star}}} - g_{\mathcal{F}_{R}}}_{n}^{2} \\
    \leq\,&
    R_{n}(\alpha_{t^{\star}})
    - R_{n}(\alpha'(D_{n}))
    + R_{n}(\alpha'(D_{n}))
    -R_{n}(g_{\mathcal{F}_{R}})
    + \inlinenorm{g_{\alpha_{t^{\star}}} - g_{\alpha'(D_{n})}}_{n}^{2}
    + \inlinenorm{g_{\alpha'(D_{n})} - g_{\mathcal{F}_{R}}}_{n}^{2} \\
    \leq\,&\varepsilon.
  \end{align*}
   Thus, the triple $(P, \mathcal{F}_{R}, g_{\alpha_{t^{\star}}})$
   satisfies the offset condition with parameters
   $(c=1/2, \varepsilon)$. In addition, by
   Equation~\eqref{eq:rkhs-norm-bound-bregman-divergence}
   we have
   \begin{align*}
    \norm{g_{\alpha_{t^{\star}}} - g_{\mathcal{F}_{R}}}_{\mathcal{H}}^{2}
    &\leq
    2\norm{g_{\alpha_{t^{\star}}} - g_{\alpha'(D_{n})}}_{\mathcal{H}}^{2}
    + 2\norm{g_{\alpha'(D_{n})} - g_{\mathcal{F}_{R}}}_{\mathcal{H}}^{2} \\
    &\leq
     2 D_{\psi}(\alpha'(D_{n}), \alpha_{t^{\star}}) + 8R^{2} \\
    &\leq \underbrace{10R^{2} + 2(LR + M)^{2}}_{\text{denote by }(R')^{2}}.
   \end{align*}
   Hence $g_{\alpha_{t^{\star}}} - g_{\mathcal{F}_{R}} \in \mathcal{F}_{R'}$,
   where $R'$ depends only on $R, L$ and $M$.\footnote{
     A smaller step-size $\eta$ can make $R'$ depend only on $R$. See the
     discussion following the statement of
     Theorem~\ref{thm:smooth-discretization}.
   }
   Applying \citep[Theorem 3]{liang2015learning}, as stated in
   Equation~\eqref{eq:bound-in-expectation-no-assumptions},
   there exist constants $c_{1}', c_{2}', c_{3}', c_{4}' > 0$, that depend only on
   $M, L$ and $R$, such that
   $$
    \ex{\mathcal{E}(g_{\alpha_{t^{\star}}}, \mathcal{F}_{R})}
    \leq
    c_{1}'\ex{\mathfrak{R}_{n}(\mathcal{F}_{R'}, c_{2}')}
    + \varepsilon
    \leq
    c_{3}'\ex{\mathfrak{R}_{n}(\mathcal{F}_{R}, c_{4}')}
    + \varepsilon.
   $$
   The result follows by taking
   $\varepsilon = c_{3}'\ex{\mathfrak{R}_{n}(\mathcal{F}_{R},
   c_{4}')}$.
\end{proof}

\subsection{In-Sample Linear Prediction Under \texorpdfstring{$\ell_{1}$}{l1}
  Constraints}
\label{section:application-l1}
Let $Z \in \R^{n \times d}$ be a fixed-design matrix such that the $\ell_{2}$
norms of columns of $Z / \sqrt{n}$ are bounded
by some constant $\kappa$.
Assume a well-specified model, i.e., the existence of a vector $\alpha'$
such that the observations $y \in \R^{n}$ follow the distribution
$y = Z\alpha' + \xi$, where $\xi$ is a vector with i.i.d.\ zero-mean
$\sigma^{2}$-subGaussian components.
We aim to find a vector $\alpha \in \R^{d}$ that achieves a small in-sample
prediction error defined as $\frac{1}{n}\norm{Z\alpha - Z\alpha'}_{2}^{2}$.

A candidate implicit regularization based algorithm,
known to be minimax optimal for sparse recovery under restricted
isometry assumption \citep{zhao2019implicit, vaskevicius2019implicit},
is defined as follows. Let $\alpha_{t} \in \R^{d}$ denote the
iterate obtained at time $t$,
let $\odot$ denote the Hadamard product, and
let $\mathbb{1}$ denote a vector with all entries equal to one.
Consider the parametrization
$\alpha_{t} = u_{t} \odot u_{t} - v_{t} \odot v_{t}$ where
$u_{t}, v_{t} \in \R^{d}$.
Instead of running gradient descent directly on $\alpha_{t}$, the algorithm
considered in the works \citep{zhao2019implicit, vaskevicius2019implicit}
is defined by running gradient descent updates on the
concatenated parameter vector
$(u, v)$, yielding the following updates (where $\gamma \in \R$):
\begin{align*}
  u_{0} = v_{0} = \sqrt{\gamma/2} \cdot \mathbb{1},
  &\quad
  \alpha_{t} = u_{t} \odot u_{t} - v_{t} \odot v_{t}, \\
  u_{t+1} = u_{t} \odot (\mathbb{1} - 2\eta \nabla R_{n}(\alpha_{t})),
  &\quad
  v_{t+1} = v_{t} \odot (\mathbb{1} + 2\eta \nabla R_{n}(\alpha_{t})).
\end{align*}
We remark that the above updates were also studied in
\citep{woodworth2019kernel}, albeit with a focus on how the initialization scale
affects the gradient descent solution obtained at convergence.
Noting that $1 + x \approx e^{x}$ for small $x$, we can approximate the above
updates (with the step-size $\eta$ rescaled by a constant factor) by
the unconstrained $\text{EG}\pm$ algorithm
\citep{kivinen1997exponentiated} whose updates are given by
\begin{align*}
  \begin{split}
    \alpha_{0}^{+} = \alpha_{0}^{-} = (\gamma/2)\mathbb{1},
    &\quad
    \alpha_{t} = \alpha_{t}^{+} - \alpha_{t}^{-}, \\
    \alpha_{t+1}^{+} = \alpha_{t}^{+}\odot\exp(-\eta \nabla R_{n}(\alpha_{t})),
    &\quad
    \alpha_{t+1}^{-} = \alpha_{t+1}^{-}\odot\exp(\eta \nabla R_{n}(\alpha_{t})).
  \end{split}
\end{align*}
It was shown in \citep{ghai2019exponentiated} that the above updates
correspond to running unconstrained mirror descent initialized at $0$ with
the mirror map given by
$$
  \psi(\alpha)
  = \phi_{\gamma}(\alpha)
  = \sum_{i=1}^{d}
  \left(
    \alpha_{i}\operatorname{arcsinh}
    \left(\alpha_{i}/\gamma\right)
    - \sqrt{\alpha_{i}^{2} + \gamma^{2}}
  \right).
$$
In the rest of the section, we denote $\psi$ by $\phi_{\gamma}$
to make the dependence on $\gamma$ explicit.
We consider running mirror descent with the hyperbolic entropy mirror map
$\phi_{\gamma}$ with any $0 < \gamma \leq (\inlinenorm{\alpha'}_{1} \wedge 1)/(3e^{2}d)$
and with any step-size $\eta$ that satisfies
  $
    0 \leq \eta \leq
    \frac{1}{24\kappa^{2}\inlinenorm{\alpha'}_{1}\log(3\gamma^{-1})}
    \wedge
    \frac{\inlinenorm{\alpha'}_{1}}{2\sigma^{2}}$.
The theorem below yields minimax-optimal rates \cite{raskutti2011minimax}
for the in-sample prediction error up to the multiplicative factor $\log(3\gamma^{-1})$.

\begin{theorem}
  \label{thm:l1-prediction-almost-optimal}
  Consider the setup described above. There exists a data-dependent
  stopping time $t^{\star} \leq \sqrt{n}/(\eta \cdot 3\kappa \sigma \sqrt{\log d})$
  such that
  with probability at least $1 - 2e^{-nc} - \frac{1}{8d^{3}}$,
  where $c$ is an absolute constant, we have
  $$
    \frac{1}{n}\norm{Z\alpha_{t^{\star}} - Z\alpha'}_{2}^{2}
    \leq 36
    \cdot \frac{\kappa\norm{\alpha'}_{1}\sigma \sqrt{\log d}}{\sqrt{n}}
    \cdot \log(3\gamma^{-1}).
  $$
\end{theorem}

Before proving the above theorem, we state two lemmas,
which relate the Bregman divergence induced by the mirror map $\phi_{\gamma}$
to the geometry induced by the $\ell_{1}$ norm.
We prove both lemmas at the end of this section.

\begin{lemma}
  \label{lemma:l1-proof-bounds-at-zero}
  For any $0 < \gamma < (\norm{\alpha'}_{1} \wedge 1)/(3e^{2}d)$
  we have
  \begin{equation*}
    \norm{\alpha'}_{1}
    \leq
    D_{\phi_{\gamma}}(\alpha', 0)
    \leq \norm{\alpha'}_{1}\log(3\gamma^{-1}).
  \end{equation*}
\end{lemma}

Denote by $\mathcal{B}_{R} = \{ w \in \R^{d} \mid \norm{w}_{1} \leq R \}$
an $\ell_{1}$ ball of radius $R$.
The following lemma will be applied to show that before stopping,
the mirror descent iterates $(\alpha_{t})_{t\geq 0}$ stay
inside an $\ell_{1}$ ball with radius at most
$6\norm{\alpha'}_{1}\log(3\gamma^{-1})$.

\begin{lemma}
  \label{lemma:l1-final-bound}
  For any $\alpha' \in \R^{d}$ and any
  $0 < \gamma < (\norm{\alpha'}_{1} \wedge 1)/(3e^{2}d)$ we have
  $$
    \left\{
      \alpha \in \R^{d} :
      D_{\phi_{\gamma}}(\alpha', \alpha) \leq 2D_{\phi_{\gamma}}(\alpha', 0)
    \right\}
    \subseteq
    \mathcal{B}_{6\norm{\alpha'}_{1}\log(3\gamma^{-1})}.
  $$
\end{lemma}

We are now ready to prove Theorem~\ref{thm:l1-prediction-almost-optimal}.
We remark that since the slow rate
$n^{-1/2}$ is minimax optimal \citep{raskutti2011minimax} in the setting
considered in Theorem~\ref{thm:l1-prediction-almost-optimal}, the
localization effect provided by offset complexities is not needed in this
example. However, we can apply Theorem~\ref{thm:smooth-discretization}
together with the \emph{basic inequality} proof technique, as demonstrated
in the proof below.

\begin{proof}[Proof of Theorem~\ref{thm:l1-prediction-almost-optimal}]
  First note that since the $\ell_{2}$
  norms of the columns of $Z/\sqrt{n}$ are bounded by $\kappa$,
  the empirical loss function $R_{n}$ is $2\kappa^{2}$-smooth
  with respect to the $\ell_{1}$ norm.
  Let
  $$
    R^{\star} = 6\norm{\alpha'}_{1}\log(3\gamma^{-1}).
  $$
  As shown in \citep[Lemma 4]{ghai2019exponentiated}, $\phi_{\gamma}$ is
  also $\rho=(2R^{\star})^{-1}$-strongly convex with
  respect to the $\ell_{1}$ norm on $\mathcal{B}_{R^{\star}}$.
  Thus, we set the smoothness parameter $\beta=2\kappa^{2}$ and
  the strong convexity parameter $\rho=(2R^{\star})^{-1}$.

  Condition on the event $A_{1} = \{R_{n}(\alpha') \leq
    2\sigma^{2}\}$. Since the noise random variables are
    $\sigma^{2}$-subGaussian, by sub-Exponential concentration we have
    $\pr{A_{1}} \geq 1 - 2e^{-nc}$ where
  $c$ is an absolute constant independent of
  any problem parameters. (cf.\ \citep[Section 5.2.4]{vershynin2010introduction}).
  By Theorem~\ref{thm:smooth-discretization},
  Lemma~\ref{lemma:l1-proof-bounds-at-zero} and $R_{n}(\alpha') \leq
  2\sigma^{2}$, it is hence enough to set
  $$
    \eta
    \leq \frac{1}{4\kappa^{2}R^{\star}} \wedge
    \frac{\norm{\alpha'}_{1}}{2\sigma^{2}}
    \leq
    \frac{\rho}{2\beta} \wedge
    \frac{D_{\psi}(\alpha', 0)}{\mathcal{L}(\alpha')}
  $$
  so that there exists a stopping time
  $$
    t^{\star}
    \leq
    \frac{2D_{\phi_{\gamma}}(\alpha', 0)}{\eta \varepsilon}
    \leq \frac{R^{\star}}{3\eta\varepsilon}
  $$
  such that for all $t \leq t^{\star}$ it holds that
  \begin{equation}
    \alpha_{t} \in
    \{\alpha \in \R^{d} : D_{\psi}(\alpha', \alpha) \leq
      D_{\psi}(\alpha', 0) + \eta R_{n}(\alpha') \}
      \subseteq
      \mathcal{B}_{R^{\star}}
      \quad\text{(cf.\ Lemma~\ref{lemma:l1-final-bound})}
  \end{equation}
  and also, such that the following inequality holds:
  \begin{equation}
      R_{n}(\alpha_{t^{\star}}) - R_{n}(\alpha')
      + \frac{1}{n}\norm{Z\alpha_{t^{\star}} - Z\alpha'}_{2}^{2} \leq \varepsilon.
  \end{equation}
  Rearranging the above inequality, as is typically done via the
  basic inequality proof technique
  (see, for example, \citep[Theorem 7.20]{wainwright2019high})
  we obtain
  \begin{align*}
    \frac{1}{n}\norm{Z\alpha_{t^{\star}} - Z\alpha'}_{2}^{2}
    \leq
    \ip{\frac{1}{n}Z^{\mathsf{T}}\xi}{\alpha_{t^{\star}} - \alpha'} + \varepsilon
    \leq
    \frac{1}{n}\norm{Z^{\mathsf{T}}\xi}_{\infty}
    \norm{{\alpha_{t^{\star}} - \alpha'}}_{1} + \varepsilon.
  \end{align*}

  Define the event $A_{2} = \{\frac{1}{n}\inlinenorm{Z^{\mathsf{T}}\xi}_{\infty}
                 \leq 4\kappa\sigma\sqrt{\log d}/\sqrt{n}\}$.
  Since the $\ell_{2}$ norms of the columns of $Z/\sqrt{n}$ are bounded by $\kappa$
  and since the noise vector $\xi$ consists of independent $\sigma^{2}$
  sub-Gaussian random variables, $\mathbb{P}(A_{2}) \geq 1 - \frac{1}{8d^{3}}$
  by standard sub-Gaussian concentration.

  By the union bound, the events $A_{1}$ and $A_{2}$ happen simultaneously with
  probability at least $1 - 2e^{-nc_{6}} - \frac{1}{8d^{3}}$.
  Setting $\varepsilon = R^{\star}\kappa\sigma\sqrt{\log d}/\sqrt{n}$ concludes our proof.
\end{proof}

%---------------------- Proof of Lemma (Bounds at Zero) -----------------------
%------------------------------------------------------------------------------
\begin{proof}[Proof of Lemma~\ref{lemma:l1-proof-bounds-at-zero}]
  The upper-bound is shown in
  \citep[Section 3]{ghai2019exponentiated}.
  We proceed as follows to prove the lower-bound:
  \begin{align}
    D_{\phi_{\gamma}}(\alpha', \alpha)
    &=
    \sum_{i=1}^{d}
    \left[
      \alpha'_{i}\left(
      \operatorname{arcsinh}\left(\frac{\alpha'_{i}}{\gamma}\right)
      -
      \operatorname{arcsinh}\left(\frac{\alpha_{i}}{\gamma}\right)
      \right)
      - \sqrt{\left(\alpha'_{i}\right)^{2} + \gamma^{2}}
      + \sqrt{\alpha_{i}^{2} + \gamma^{2}}
    \right] \notag \\
    &\geq
    \sum_{i=1}^{d}
      \alpha'_{i}\left(
      \operatorname{arcsinh}\left(\frac{\alpha'_{i}}{\gamma}\right)
      -
      \operatorname{arcsinh}\left(\frac{\alpha_{i}}{\gamma}\right)
      \right)
      - 2\norm{\alpha'}_{1}
      + \norm{\alpha}_{1}
    \notag \\
    &\geq
    \sum_{i=1}^{d}
      \alpha'_{i}
      \operatorname{arcsinh}\left(\frac{\alpha'_{i}}{\gamma}\right)
      -
    \norm{\alpha'}_{1}
    \operatorname{arcsinh}\left(\frac{\norm{\alpha}_{1}}{\gamma}\right)
    - 2\norm{\alpha'}_{1}
    + \norm{\alpha}_{1}
    \notag \\
    &=
    \sum_{i=1}^{d}
      \abs{\alpha'_{i}}
      \log\frac{\abs{\alpha'_{i}} + \sqrt{\left(\alpha'_{i}\right)^{2} +
                \gamma^{2}}}
               {\gamma}
      -
    \norm{\alpha'}_{1}
    \operatorname{arcsinh}\left(\frac{\norm{\alpha}_{1}}{\gamma}\right)
    - 2\norm{\alpha'}_{1}
    + \norm{\alpha}_{1}
    \notag \\
    &\geq
    \norm{\alpha'}_{1}\log\frac{\norm{\alpha'_{1}}}{d\gamma}
    - \norm{\alpha'}_{1}
    \operatorname{arcsinh}\left(\frac{\norm{\alpha}_{1}}{\gamma}\right)
    - 2\norm{\alpha'}_{1}
    + \norm{\alpha}_{1} \notag \\
    &=
    \norm{\alpha'}_{1}\log\frac{\norm{\alpha'_{1}}}{e^{2}d\gamma}
    - \norm{\alpha'}_{1}
    \operatorname{arcsinh}\left(\frac{\norm{\alpha}_{1}}{\gamma}\right)
    + \norm{\alpha}_{1}, \label{eq:starting-point-bregman-lower-bound}
  \end{align}
  where the penultimate line follows via an application of the log sum
  inequality. The result follows by plugging in $\norm{\alpha}_{1} = 0$ and
  using $\gamma \leq \norm{\alpha'}_{1}/(e^3d)$.
\end{proof}

%-------------------------- Proof of Final l1 Bound ---------------------------
%------------------------------------------------------------------------------
\begin{proof}[Proof of Lemma~\ref{lemma:l1-final-bound}]
  Note that for any $x > \gamma > 0$, we have
  $\operatorname{arcsinh}(x/\gamma) \leq \log(3x/\gamma)$.
  Hence, continuing from Equation~\eqref{eq:starting-point-bregman-lower-bound}
  we have
  \begin{align*}
    &\norm{\alpha}_{1} > \norm{\alpha'}_{1} \\
    \implies&
    \norm{\alpha}_{1}
    \leq
    D_{\phi_{\gamma}}(\alpha', \alpha) +
    \norm{\alpha'}_{1}
    \left(
      \log\left(3e^{2}d\right)
      + \log \frac{\norm{\alpha}_{1}}{\norm{\alpha'}_{1}}
    \right) \\
    &\hphantom{\norm{\alpha}_{1}\,}
    \leq
    D_{\phi_{\gamma}}(\alpha', \alpha) +
    \norm{\alpha'}_{1}
    \left(
      \log\frac{1}{\gamma}
      + \frac{1}{2}\frac{\norm{\alpha}_{1}}{\norm{\alpha'_{1}}}
    \right) \\
    \implies&
    \norm{\alpha}_{1}
    \leq
    2D_{\phi_{\gamma}}(\alpha', \alpha) + 2\norm{\alpha'}\log\frac{1}{\gamma}.
\end{align*}
The result follows by applying the upper-bound proved in
Lemma~\ref{lemma:l1-proof-bounds-at-zero}, namely,
$D_{\phi_{\gamma}}(\alpha', \alpha) \leq 2D_{\phi_{\gamma}}(\alpha', 0)
\leq 2\inlinenorm{\alpha'}_{1}\log(3\gamma^{-1})$.
\end{proof}

\subsection{Statistical Guarantees Along the Optimization Path}
\label{section:application-optimization-path}

Theorem~\ref{thm:continuous-time} immediately implies that along its
optimization path, continuous-time mirror descent satisfies excess risk
guarantees that one would obtain for a series of ERM solutions over
a corresponding set of \emph{explicitly constrained} convex and bounded
problems.

Fix any radius $R > 0$, let $M = \inlinenorm{Y}_{L_{\infty}(P)}$ and suppose that
$\sup_{g \in \mathcal{F}(\alpha_{0}, R)} \abs{g}_{\infty} \leq B$ for
some constant $B > 0$.
For simplicity, in the proof below, we assume that the coordinate system
represented by the parametrization $\alpha \in \R^{m}$ is independent of the data.
A data dependent coordinate system can be handled via the same technique
used to prove Theorem~\ref{thm:nonparametric-regression}, thus we note
that the result stated in Theorem~\ref{thm:implicit-vs-explicit}
also holds in the kernel regime.
For smooth loss functions and sufficiently small
step-sizes, Theorem~\ref{thm:smooth-discretization} can be applied to
derive a corresponding result in discrete time.

\begin{theorem}
  \label{thm:implicit-vs-explicit}
  Fix any $\alpha_{0} \in \R^{m}$, $R > 0$ and let
  $
    \mathcal{F}(\alpha_{0}, R) = \{
      g_{\alpha} : D_{\psi}(\alpha, \alpha_{0}) \leq R
    \}.
	 $
  For any $\varepsilon > 0$, there exists a data-dependent stopping time
  $t^{\star} \leq 2R/\varepsilon$
  such that for some $c_{1}, c_{2} > 0$, depending only on the
  boundedness constants $B$ and $M$, we have
  \[
    \ex{\mathcal{E}(g_{\alpha_{t^{\star}}}, \mathcal{F}(\alpha_{0}, R))}
    \leq
    c_{1}\ex{
      \mathfrak{R}_{n}(\mathcal{F}(\alpha_{0}, R) -
      g_{\mathcal{F}(\alpha_{0}, R)}, c_{2})
    }
    + \varepsilon.
 \]
\end{theorem}

\begin{proof}[Proof of Theorem~\ref{thm:implicit-vs-explicit}]
  Let $\alpha'$ be the parameter corresponding to the function
  $g_{\mathcal{F}(\alpha_0, R)}$.
  Then, by Theorem~\ref{thm:continuous-time}
  there exists a data-dependent stopping time $t^{\star} \leq 2D_{\psi}(\alpha',
  \alpha_{0})/\varepsilon \leq 2R/\varepsilon$ such that the triple
  $(P, \mathcal{F}(\alpha_{0}, R), g_{\alpha_{t^{\star}}})$ satisfies the
  offset condition (cf.\ Definition~\ref{definition:offset-condition})
  with parameters $(c=1, \varepsilon)$
  and $g_{\alpha_{t^{\star}}} \in \mathcal{F}(\alpha_{0}, R)$.
  Thus, by \citep[Theorem 3]{liang2015learning}, as stated in
  Equation~\eqref{eq:bound-in-expectation-no-assumptions}, there exist
  constants $c_{1}, c_{2}$ that depend only on $B$ and $M$ such that
  $\ex{\mathcal{E}(g_{\alpha_{t^{\star}}}, \mathcal{F}(\alpha_{0}, R))} \leq c_{1}\ex{
    \mathfrak{R}_{n}(\mathcal{F}(\alpha_{0}, R) - g_{\mathcal{F}(\alpha_{0}, R)}, c_{2})
  } + \varepsilon$.
\end{proof}

Through the lens of the offset complexities, similar excess risk guarantees hold
for any ERM estimator over the function space $\mathcal{F}(\alpha_{0}, R)$, which we denote by
$\widehat{g}_{\mathcal{F}(\alpha_{0}, R)}$.
To see why, note that since Bregman divergences are convex in the first argument,
the class $\mathcal{F}(\alpha_{0}, R)$ is convex.
Thus, by \citep[Lemma 1]{liang2015learning}, the triple
$(P, \mathcal{F}(\alpha_{0}, R), \widehat{g}_{\mathcal{F}(\alpha_{0}, R)})$
satisfies
the offset condition (cf.\ Definition~\ref{definition:offset-condition})
with parameters $(c=1, \varepsilon=0)$.
Hence, with the same choice of $c_{1}$ and $c_{2}$ as in
Theorem~\ref{thm:implicit-vs-explicit}, it holds that
$$
  \ex{\mathcal{E}(
    \widehat{g}_{\mathcal{F}(\alpha_{0}, R)}, \mathcal{F}
  )}
  \leq c_{1}\ex{
    \mathfrak{R}_{n}(\mathcal{F}(\alpha_{0}, R) - g_{\mathcal{F}(\alpha_{0}, R)}, c_{2})
  }.
$$
In particular, Theorem~\ref{thm:implicit-vs-explicit} shows, that along the
optimization path $(g_{\alpha_{t}})_{t \geq 0}$, the mirror descent iterates
satisfy almost identical excess risk guarantees that also hold for
the regularization path $(\widehat{g}_{\mathcal{F}(\alpha_{0}, R)})_{R \geq 0}$ obtained by
ERM solutions over classes $\mathcal{F}(\alpha_{0}, R)$ with varying radius~$R$.

The above observation is related to some of the recent work in the implicit
regularization literature, which has sought to provide statistical
guarantees on solutions along the optimization path of gradient descent and
mirror descent. The work \citep{suggala2018connecting} establishes
that the optimization paths and the regularization paths of \emph{corresponding} regularized
problems, when suitably aligned via some mapping between the number of
mirror descent iterations and the regularization parameter of the penalized
problem,
are point-wise close. This allows them to port
existing results on regularized optimization to early-stopped descent
algorithms; in contrast, our result proved in
Theorem~\ref{thm:implicit-vs-explicit} shows that the excess risk of
solutions along the optimization path of mirror descent can
be directly bounded by the offset Rademacher complexity. We also do not require
the loss function to be strongly convex as a function of parameter $\alpha$.
The work \citep{ali2019continuous} studies the
optimization path of continuous time gradient descent for linear regression,
which can be computed analytically, and show that the solution at time $t$ has
risk at most $1.69$ times the risk of the \emph{ridge} solution with $\lambda =
1/t$. In contrast to our results, their results are derived under the
well-specified model and their out-of-sample analysis is
performed via a certain Bayesian averaging that is not needed in our work.

\section{Future Directions}

Our work provides a simple and transparent framework for
simultaneously analyzing statistical and computational properties of iterates
traced by a family of mirror
descent algorithms applied to the i.i.d.\ batch statistical learning setting.
Among the research directions that would yield additional computational savings
are extensions of our results to stochastic and accelerated frameworks,
where connections between early stopping and localized complexity measures
are yet to be established, even in the restricted setting of
Euclidean~gradient~descent~updates.

Beyond the computational savings, our main results reveal a curious property of
mirror descent. For an unknown parameter of interest denoted by $\alpha'$, the
statistical complexity of an appropriately stopped mirror descent iterate is
given by the offset
complexity of the class $\{g_{\alpha} - g_{\alpha'} : D_{\psi}(\alpha',
  \alpha) \leq D_{\psi}(\alpha', \alpha_{0})\}$. Thus, $g_{\alpha}$ is
 \emph{implicitly constrained} to lie in a \emph{possibly non-convex} Bregman ball
 \emph{centered at} the unknown $\alpha'$ with \emph{unknown radius}
 $D_{\psi}(\alpha', \alpha_{0})$.
 Therefore, in general, solutions traced by mirror descent iterates
cannot be practically expressed as solutions of \emph{explicitly constrained}
optimization problems. Consequently, early-stopped mirror descent can potentially
solve problems that cannot be tractably solved by the means of explicit
regularization. This observation necessitates further investigation.

\subsection*{Acknowledgments}

Tomas Va\v{s}kevi\v{c}ius is supported by the EPSRC and MRC through the
OxWaSP CDT programme (EP/L016710/1).
Varun Kanade and Patrick Rebeschini are supported in part by the Alan Turing
Institute under the EPSRC grant EP/N510129/1.

%-------------------------------- Bibliography --------------------------------
%------------------------------------------------------------------------------
\bibliographystyle{plainnat}
\bibliography{main}

%---------------------------------- Appendix ----------------------------------
%------------------------------------------------------------------------------
%\clearpage
\appendix

\section{On the \texorpdfstring{$\varepsilon$}{Epsilon} Term in the Offset
           Condition}
\label{appendix:on-epsilon-term}

Suppose that an estimator $\widehat{g}$ that always returns a function
from some set $\mathcal{G}$ and the triple $(P, \mathcal{F}, \widehat{g})$
satisfies the offset condition (cf.\ Definition~\ref{definition:offset-condition})
with parameters $(\varepsilon, c)$.
Then, following along the lines of \citet[Corollary 2]{liang2015learning}
we obtain
\begin{align*}
  &\phantom{{}={}}R(\widehat{g}) - R(g_{\mathcal{F}}) \\
  &=R(\widehat{g}) - R(g_{\mathcal{F}})
  - (R_{n}(\widehat{g}) - R_{n}(g_{\mathcal{F}}))
  + (R_{n}(\widehat{g}) - R_{n}(g_{\mathcal{F}})) \\
  &\leq
   R(\widehat{g}) - R(g_{\mathcal{F}})
   - (R_{n}(\widehat{g}) - R_{n}(g_{\mathcal{F}}))
   - c\norm{\widehat{g} - g_{\mathcal{F}}}_{n}^{2} + \varepsilon \\
  &=
     \ex{2(\widehat{g}(X) - g_{\mathcal{F}}(X))
             (g_{\mathcal{F}}(X) - Y)}
     - \frac{1}{n}\sum_{i=1}^{n}2(\widehat{g}(x_{i}) - g_{\mathcal{F}}(x_{i}))
                            (g_{\mathcal{F}}(x_{i}) - y_{i}) \\
  &\phantom{{}={}}+ \norm{\widehat{g} - g_{\mathcal{F}}}_{P}^{2}
     - (1+c)\norm{\widehat{g} - g_{\mathcal{F}}}_{n}^{2} + \varepsilon \\
  &\leq
  \sup_{g \in \mathcal{G} - g_{\mathcal{F}}}
  \left\{
     \ex{2g(X) (g_{\mathcal{F}}(X) - Y)}
     - \frac{1}{n}\sum_{i=1}^{n}2g(x_{i}) (g_{\mathcal{F}}(x_{i}) - y_{i})
     + \norm{g}_{P}^{2}
     - (1+c)\norm{g}_{n}^{2}
  \right\}
     + \varepsilon. \\
\end{align*}
  The penultimate term, excluding $\varepsilon$,
  is the same as the term obtained
  in \citep[Corollary 2]{liang2015learning}.
  Since the range of $\widehat{g}$ is $\mathcal{G}$, the function
  $\widehat{g} - g_{\mathcal{F}}$ is constrained to lie in the set
  $\mathcal{G} - g_{\mathcal{F}}$, which gives the term on the last
  line involving the supremum (cf.\ \citep[Equation 11]{liang2015learning}).
  From this point, offset Rademacher complexity bounds in expectation
  \citep[Theorem 3]{liang2015learning} and
  probability \citep[Theorem 4]{liang2015learning} are obtained via
  symmetrization of the above term in expectation and
  probability respectively.

\section{Table of Notation}
\label{appendix:table-of-notation}

\begin{table}[h]
  \caption{Table of notation}
  \label{table:notation}
  \centering
  \begin{tabular}{ll}
    \toprule
    Symbol
    & Description \\
    \midrule
    $n$
    & The number of data points. \\

    $P$
    & The data generating distribution. \\

    $(x_{i}, y_{i})$
    & The $i$\textsuperscript{\it th} data point sampled independently from
      the distribution $P$. \\

    $D_{n}$
    & A collection of $n$ data points sampled i.i.d.\ from $P$. \\

    $R(g)$
    & The population risk of a function $g$ defined as
      $\exd{(X, Y)\sim P}{(g(X) - Y)^{2}}$. \\

    $R_{n}(g)$
    & The empirical risk of a function $g$ defined as
      $\frac{1}{n}\sum_{i=1}^{n}(g(x_{i}) - y_{i})^{2}$. \\

    $\widehat{g}$
    & An estimator, which maps datasets $D_{n}$ to some set of functions $\mathcal{G}$.
    \\

    $\mathcal{G}$
    & The set of possible values of functions, that some estimator
      $\widehat{g}$ can select.\\

    $\mathcal{F}$
    & Some generic class of functions. \\

    $\mathcal{E}(\widehat{g}, \mathcal{F})$
    & The excess risk $R(\widehat{g}) - \inf_{g \in \mathcal{F}}R(g)$
      of an estimator $\widehat{g}$ with respect to $\mathcal{F}$. \\

    $g_{\mathcal{F}}$
    & A function $g \in \mathcal{F}$ such that
      $R(g) = \inf_{g\in\mathcal{F}}R(g)$. \\

    $\inlinenorm{g - f}_{P}^{2}$
    & Population $\ell_{2}$ distance between $g$ and $f$ defined as
      $\ex{(g(X) - f(X))^{2}}$ \\

    $\inlinenorm{g - f}_{n}^{2}$
    & Empirical $\ell_{2}$ distance between $g$ and $f$ defined as
      $\frac{1}{n}\sum_{i=1}^{n}(g(x_{i}) - f(x_{i}))^{2}$. \\

    $\mathfrak{R}_{n}(\mathcal{G}, c)$
    & The offset Rademacher complexity of
      $\mathcal{G}$ (cf.\ Equation~\eqref{eq:offset-complexity-definition}). \\

    $m$
    & The dimensionality of the parameter space. \\

    $g_{\alpha}$
    & A function parameterized by $\alpha \in \R^{m}$. \\

    $Z \in \R^{n \times m}$
    &  A matrix such that conditionally on $D_{n}$,
       $g_{\alpha}(x_{i}) = (Z\alpha)_{i}$ for any
       $\alpha \in \R^{m}$. \\

    $R_{n}(\alpha)$
    & A shorthand notation for
      $R_{n}(g_{\alpha}) = \frac{1}{n}\inlinenorm{Z\alpha - y}^{2}_{2}$. \\

    $\psi$
    & A mirror map. \\

    $D_{\psi}$
    & Bregman divergence induced by the mirror map $\psi$. \\

    $\alpha_{0}$
    & The initialization point of the mirror descent iterates. \\

    $\alpha_{t}$
    & The mirror descent iterate at time $t$.  \\

    $\alpha'$
    & An arbitrarily chosen reference point. \\

    $\delta_{t}$
    & A shorthand notation for $R_{n}(\alpha_{t}) - R_{n}(\alpha')$. \\

    $r_{t}$
    & A shorthand notation for $\inlinenorm{g_{\alpha_{t}} - g_{\alpha'}}_{n}^{2}$.
  \end{tabular}
\end{table}

\end{document}